%% file: main.tex
\documentclass[10pt,twocolumn,letterpaper]{article}

\usepackage{cvpr}              

\input{preamble}

\definecolor{cvprblue}{rgb}{0.21,0.49,0.74}
\usepackage[pagebackref,breaklinks,colorlinks,allcolors=cvprblue]{hyperref}

\def\paperID{21908} 
\def\confName{CVPR}
\def\confYear{2026}

\title{Spatiotemporal Pyramid Flow Matching for Climate Emulation}

\author{
\textbf{Jeremy A. Irvin}$^{1}$\thanks{Correspondence to: \href{jirvin16@cs.stanford.edu}{\texttt{jirvin16@cs.stanford.edu}}}\,,
\textbf{Jiaqi Han}$^{1}$,
\textbf{Zikui Wang}$^{1}$,
\textbf{Abdulaziz Alharbi}$^{1}$,
\textbf{Yufei Zhao}$^{1}$,\\
\textbf{Nomin-Erdene Bayarsaikhan}$^{1}$,
\textbf{Daniele Visioni}$^{2}$,
\textbf{Andrew Y. Ng}$^{1}$,
\textbf{Duncan Watson-Parris}$^{3}$\\[4pt]
$^{1}$Stanford University \qquad
$^{2}$Cornell University \qquad
$^{3}$University of California, San Diego
}

\begin{document}
\maketitle

\input{sec/0_abstract}
\input{sec/1_intro}
\input{sec/2_methods}

\input{sec/3_data}

\input{sec/4_experiments}

\input{sec/5_discussion}

{
    \small
    \bibliographystyle{ieeenat_fullname}
    \bibliography{main}
}

\input{sec/6_suppl}

\end{document}

%% file: preamble.tex
\usepackage{amsmath}
\usepackage{graphicx}
\usepackage{multirow}
\usepackage{enumitem}
\usepackage{pifont}
\usepackage{titletoc}

\usepackage[table]{xcolor}

\definecolor{pastelgreen}{HTML}{DFF0D8}

\newcommand{\cmark}{\textcolor{green!60!black}{\ding{51}}}%
\newcommand{\xmark}{\textcolor{red!70!black}{\ding{55}}}%

\newcommand{\lowerp}[1]{\raisebox{-0.2ex}[0pt][0pt]{#1}}
\newcommand{\raisep}[1]{\raisebox{0.2ex}[0pt][0pt]{#1}}

%% file: sec/0_abstract.tex
\begin{abstract}
Generative models have the potential to transform the way we emulate Earth’s changing climate. Previous generative approaches rely on weather-scale autoregression for climate emulation, but this is inherently slow for long climate horizons and has yet to demonstrate stable rollouts under nonstationary forcings. Here, we introduce Spatiotemporal Pyramid Flows (SPF), a new class of flow matching approaches that model data hierarchically across spatial and temporal scales. Inspired by cascaded video models, SPF partitions the generative trajectory into a spatiotemporal pyramid, progressively increasing spatial resolution to reduce computation and coupling each stage with an associated timescale to enable direct sampling at any temporal level in the pyramid. This design, together with conditioning each stage on prescribed physical forcings (e.g., greenhouse gases or aerosols), enables efficient, parallel climate emulation at multiple timescales. On ClimateBench, SPF outperforms strong flow matching baselines and pre-trained models at yearly and monthly timescales while offering fast sampling, especially at coarser temporal levels. To scale SPF, we curate ClimateSuite, the largest collection of Earth system simulations to date, comprising over 33,000 simulation-years across ten climate models and the first dataset to include simulations of climate interventions. We find that the scaled SPF model demonstrates good generalization to held-out scenarios across climate models. Together, SPF and ClimateSuite provide a foundation for accurate, efficient, probabilistic climate emulation across temporal scales and realistic future scenarios. 
Data and code is publicly available at \href{https://github.com/stanfordmlgroup/spf}{github.com/stanfordmlgroup/spf}.
\end{abstract}

%% file: sec/1_intro.tex
\section{Introduction}
Climate models (Earth system models; ESMs) are the primary scientific instruments for quantifying how the Earth's climate will evolve under changing emissions and interventions. They resolve interacting physical processes across a wide range of spatial and temporal scales, but this fidelity comes with enormous computational cost: long rollouts (decades to centuries), ensembles for uncertainty quantification, and large design sweeps across forcings make systematic exploration prohibitive even on state-of-the-art supercomputers \cite{kay2015community,eyring2016overview,govett2024exascale}. As a consequence, scientists and policymakers are constrained to a narrow set of scenarios and limited ensemble sizes, which hinders robust assessment of regional risks and of policy-relevant choices such as emissions pathways or climate interventions \citep{o2016scenario,kravitz2015geoengineering,watsonparris2022climatebench}. These pressures have led to increasing interest in data-driven surrogates that efficiently emulate ESMs while remaining accurate, stable, and physically plausible over very long horizons \cite{watsonparris2022climatebench,wattmeyer2024ace,cachay2024sphericaldyffusion}.

Recent work has made tangible progress toward this goal, but it has relied primarily on autoregressive emulation at the weather-scale (short term) rolled out over climate-scale (long term) periods \cite{wattmeyer2024ace,watt2025ace2,cachay2024sphericaldyffusion,kochkov2024neural}. This approach inherits two central limitations for climate emulation: first, weather-scale autoregressive models are subject to small local errors that compound over time, leading to drift in long-term statistics. Although recent models have demonstrated decade-long rollouts with small climatological biases in a simplified atmospheric model with fixed forcings, they have yet to capture realistic climate trends driven by greenhouse gas emissions or aerosols \cite{wattmeyer2024ace,watt2025ace2}. Second, long-horizon rollouts remain computationally expensive due to the large number of sequential autoregressive steps; for example, generating a single 10-year trajectory with a state-of-the-art emulator takes nearly three hours \cite{cachay2024sphericaldyffusion}. This is especially prohibitive for many practical downstream uses which only require coarser timescale samples, e.g., annual indicators for integrated assessment models  and policy assessments, or monthly fields for sectoral impact studies \cite{forster2021emulators,tebaldi2025emulators}. In parallel, regression-based emulators that directly map external forcings to long-term mean climate responses offer an alternative path toward efficiency \cite{watsonparris2022climatebench,niu2024multi}. Our approach builds upon regression-based methods to capture multiple timescales within a single model and support parallel sampling of temporal sequences, leading to additional efficiency gains for long time horizons.

Beyond accuracy and efficiency, a critical requirement for climate emulation is the ability to represent uncertainty. In Earth system modeling, ensembles of simulations are used to quantify variability and assess the likelihood of climate outcomes. Similarly, probabilistic emulators should generate diverse samples from a learned distribution, allowing estimation of uncertainty and exploration of alternative trajectories under the same forcing conditions. Only recently has work begun to address this need by developing stochastic climate-scale emulators capable of sampling multiple plausible futures \cite{bouabid2024fairgp,cachay2024sphericaldyffusion,kochkov2024neural}.

Diffusion and flow-based image and video generative models offer a promising path toward emulators that combine efficiency with probabilistic generation. These models provide a flexible framework for sampling from complex multimodal data distributions and have demonstrated remarkable success on image \cite{ho2020denoising,lipman2022flow} and video generation tasks \cite{ho2022video,jin2024pyramidal}. However, most high-quality video models (i) rely on autoregression through time to model long sequences, which scales poorly \cite{chen2024diffusion} and/or (ii) compress the high-dimensional spatiotemporal data into low-dimensional latents using a strong variational autoencoder (VAE), a component that is unavailable for climate data, difficult to train, and constrains performance to the quality of the learned compression \cite{jin2024pyramidal,liu2024st,polyak2024movie,hacohen2024ltx,kong2024hunyuanvideo}. We address these challenges with an efficient parallel approach operating in pixel space, eliminating the need for a separate VAE.

A key algorithmic insight from the natural image and video domains is to cascade generation: model coarse structure first and then refine in multiple stages, thereby concentrating compute where it most affects perceptual quality \cite{ho2022cascaded}. Recent work performs this cascade within a single model by partitioning the flow into stages at different spatial resolutions, yielding faster sampling and a lower memory footprint than a full-resolution flow while allowing knowledge sharing between stages \cite{jin2024pyramidal,chen2025pixelflow}. For climate data, an analogous inductive bias exists in both space and time. Spatially, large scales organize small scales through energy and moisture transports; temporally, slow components such as forced trends and interannual variability modulate fast weather fluctuations. Rather than explicitly modeling these long-term dependencies through extended rollouts, conditioning the model on physical forcings (e.g., greenhouse gases, aerosols, or solar variability) allows the emulator to capture slow, externally driven trends without requiring access to long historical sequences, as framed in \citet{watsonparris2022climatebench}. This provides a principled path to long-horizon efficiency: learn flows that represent coarse, slow components in latent space and progressively refine them across scales to circumvent fine-step autoregression.

\begin{figure*}[t]
  \centering
  \begin{minipage}[b]{0.510\textwidth}
    \centering
    \includegraphics[width=\linewidth]{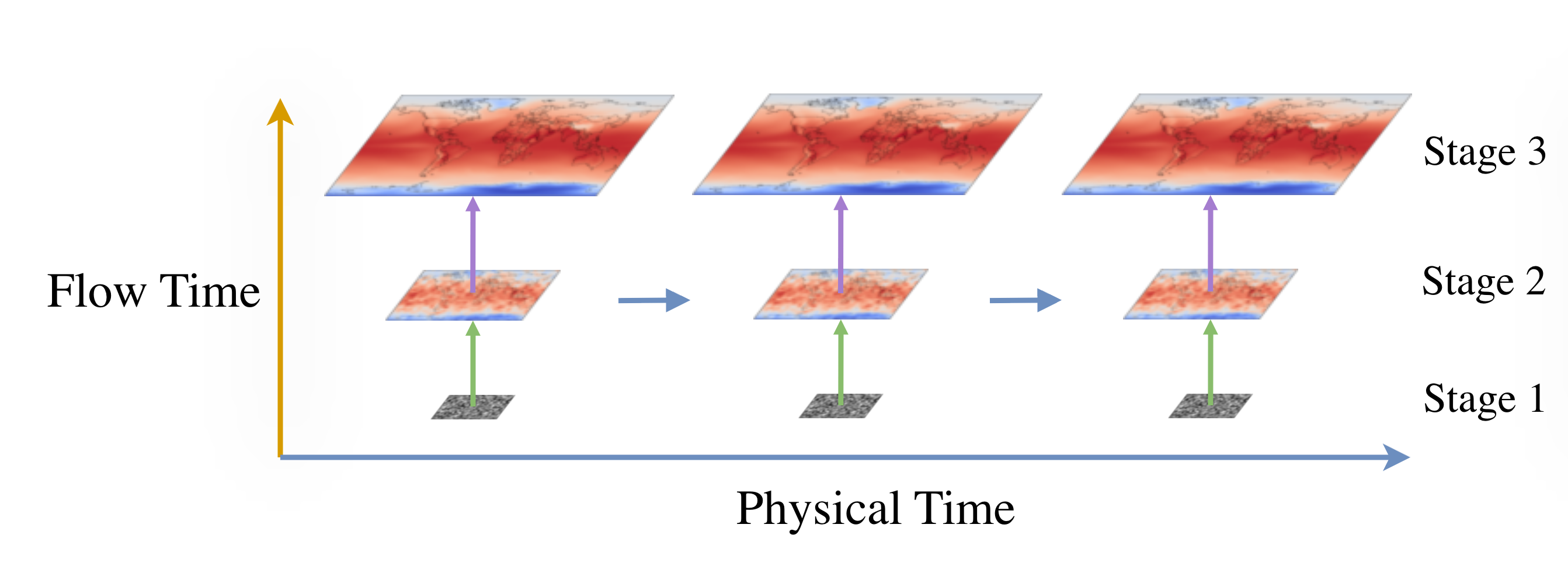}
    \vspace{-2em}
    \caption*{(a) PyramidalFlow (\citet{jin2024pyramidal})}
  \end{minipage}
  \hspace{0.5em}
  \begin{minipage}[b]{0.470\textwidth}
    \centering
    \includegraphics[width=\linewidth]{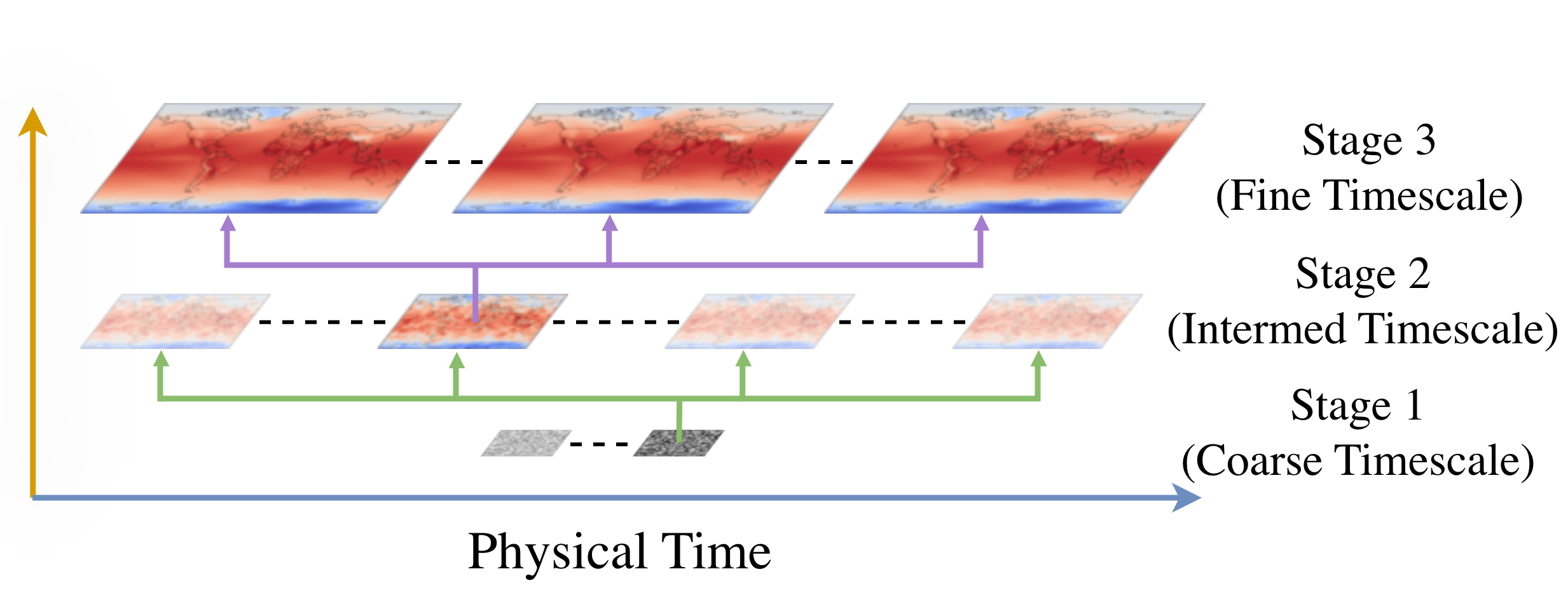}
    \vspace{-2em}
    \caption*{(b) Our Spatiotemporal Pyramid Flow (SPF)}
  \end{minipage}
  \caption{\textbf{Comparison between PyramidalFlow and our proposed approach.} (a) PyramidalFlow generates autoregressively, running one pyramid flow for every physical timestep. (b) Our method generates in parallel, running a joint flow to sample multiple timesteps at once.  To save compute for long sequences, SPF can funnel into a chosen timestep before transitioning to the next timescale. Since timescale lengths differ (e.g. 10 years for decadal or 12 months for annual), SPF supports variable temporal (and spatial) transitions between stages.}
  \label{fig:overview}
\end{figure*}

In this work, we introduce \textbf{Spatiotemporal Pyramid Flows (SPF)}, a class of flow matching models that generalize pyramid flows for efficient emulation across timescales. SPF partitions the generative trajectory into a spatiotemporal pyramid that (i) models coarse-resolution, long-horizon climate states that encode externally forced trends and low-frequency variability, (ii) conditionally refines spatial detail, and (iii) conditionally refines temporal resolution, enabling direct sampling at arbitrary future times without rolling forward step by step. By conditioning each stage on prescribed physical forcings, SPF can funnel into specific spatial or temporal windows while maintaining long-term physical coherence. Crucially, this formulation generalizes prior pyramid flow methods by allowing arbitrary resampling between stages and by supporting clean generation from any level of the hierarchy, yielding efficient sampling across timescales relevant for climate-scale downstream tasks.

To train a scaled SPF model, we curate \textbf{ClimateSuite}, the largest ML-ready climate modeling dataset to date with more than 33,000 simulation-years of data from 10 ESMs. As climate emulators are increasingly expected to support policy-relevant questions, such as responses to climate interventions, we include 39 stratospheric aerosol injection (SAI) experiments in addition to 276 non-SAI experiments.  We publicly release ClimateSuite to facilitate the development of strong and efficient climate emulators across different ESMs, emissions scenarios, and interventions.

\noindent Altogether, our contributions are as follows:
\begin{enumerate}[itemsep=0pt, parsep=6pt, topsep=0pt]
    \item We introduce a new class of flow matching approaches called Spatiotemporal Pyramid Flows (SPF) that enable efficient and accurate sampling by applying a pyramid of resolution-cascaded flows across both space and time. SPF refines noised latent trajectories conditioned on physical forcings, enabling accurate sampling of emulated climate model outputs at arbitrary future times without autoregressive simulation.
    \item We show that SPF generalizes prior pyramid flow approaches \cite{jin2024pyramidal,chen2025pixelflow}, importantly supporting arbitrary resolution resampling between stages and enabling generation of clean samples from any timescale in the pyramid.
    \item We curate and publicly release the largest dataset for climate model emulation to date, ClimateSuite, comprising more than 33,000 simulation-years of climate data spanning 276 state-of-the-art simulations from 10 ESMs and 39 SAI simulations. This substantially expands the data available for developing ML-based climate emulators.
    \item We demonstrate that SPF achieves superior accuracy and inference efficiency compared to strong deterministic baselines, pre-trained models, and flow matching approaches on ClimateBench \cite{watsonparris2022climatebench}. Training SPF on ClimateSuite leads to good generalization to emissions and intervention scenarios across climate models, and improves performance on ClimateBench after fine-tuning.
\end{enumerate}

%% file: sec/2_methods.tex
\section{Methods}
\label{sec:methods}

\subsection{Background}

\subsubsection{Diffusion and Flow Matching}
\label{sec:background-diff-fm-detailed}
Diffusion models \citep{sohl2015deep,song2019generative,ho2020denoising} and flow generative models \citep{papamakarios2021normalizing,onken2021ot,lipman2022flow,liu2022flow,yan2024perflow} transform random noise $\boldsymbol{x}_0\sim \mathcal{N}(\boldsymbol{0}, \boldsymbol{I})$ into samples from a target data distribution  $\boldsymbol{x}_1\sim q$ by learning a velocity field $\boldsymbol{v}_t$ which prescribes an ordinary differential equation (ODE):
\begin{equation*}
\frac{d\boldsymbol{x}_t}{dt}=\boldsymbol{v}_t(\boldsymbol{x}_t).
\end{equation*}
The velocity field is commonly learned using conditional flow matching, a simple simulation-free training objective to directly regress the velocity $\boldsymbol{v}_t$ with a conditional (per-sample) vector field $\boldsymbol{u}_t(\cdot | \boldsymbol{x}_1)$:
\begin{equation*}
\mathbb{E}_{t\sim\mathcal{U}[0,1],q(\boldsymbol{x}_1),p_t(\boldsymbol{x}_t|\boldsymbol{x}_1)}\| \boldsymbol{v}_t(\boldsymbol{x}_t)-\boldsymbol{u}_t(\boldsymbol{x}_t|\boldsymbol{x}_1)\|^2.
\end{equation*}

\noindent This is a tractable objective with the same optima as the marginal flow matching objective where the choice of $\boldsymbol{u}_t(\cdot | \boldsymbol{x}_1)$ uniquely determines a conditional probability path $p_t(\boldsymbol{x}_t|\boldsymbol{x}_1)$,  a time-dependent probability density function that describes how samples from the prior transform to samples from the target distribution.\\

\noindent A common and effective choice of vector field is $\boldsymbol{u}(\boldsymbol{x}_t|\boldsymbol{x}_1) = \boldsymbol{x}_1 - \boldsymbol{x}_0$, which induces a conditional probability path that linearly interpolates between noise and data:
\begin{align*}
    \boldsymbol{x}_t&=t\boldsymbol{x}_1+(1-t)\textbf{x}_0\\
    \boldsymbol{x}_t&\sim  \mathcal{N}(t\boldsymbol{x}_1, (1-t)^2\boldsymbol{I}).
\end{align*}

\subsubsection{Piecewise Flows}
\label{sec:background-piecewise}

Piecewise flows \citep{yan2024perflow} partition the flow trajectory into $K$ segments
$\{[s_k,e_k]\}_{k=0}^{K-1}$ where $0=s_{K-1}<e_{K-1}=s_{K-2}<\dots<e_{1}=s_{0}<e_{0}=1.$ Each segment is endowed with its own probability path \(p_{t}^{(k)}\) and therefore its own target vector field
\(\boldsymbol{u}^{(k)}_t\!\bigl(\boldsymbol{x}_t^{(k)}\bigr)\).

\noindent At inference time, the learned ODE on each segment is integrated in sequence, ensuring that the endpoints of each segment aligns (in distribution) with the endpoints of adjacent segments when necessary.

\subsubsection{Pyramid Flows}
\label{sec:pyramid-flows}
To reduce the computational cost of sampling images and videos, previous work has proposed constructing piecewise flows segmented by spatial resolution so only the final stage operates at the highest resolution \citep{jin2024pyramidal,chen2025pixelflow}. This design forms a spatial pyramid partitioned into $K$ stages, where stage~$k$ denoises an image at lower spatial resolution which is subsequently upsampled and input to stage~$k\!-\!1$, gradually denoising and increasing spatial resolution through the flow.\\

\noindent Formally, the flow within the $k$-th time window $[s_k, e_k]$ is:
\begin{equation*}
\resizebox{0.95\columnwidth}{!}{$
\boldsymbol{x}_t=t'\mathrm{Down}(\boldsymbol{x}_{e_k}, 2^k)+(1-t')\mathrm{Up}(\mathrm{Down}(\boldsymbol{x}_{s_k},2^{k+1})),
$}
\end{equation*}

\noindent where $t'= (t - s_k)/(e_k - s_k)$ is the rescaled timestep, $\mathrm{Down}$ is a well-defined downsampling function (e.g., bilinear interpolation), and $\mathrm{Up}$ is an upsampling function (e.g., nearest neighbor interpolation).  The conditional probability path is defined by:
{\small
\begin{align*}
  \text{End:}\quad
    \hat{\boldsymbol{x}}_{e_k} &\mid \boldsymbol{x}_1
      \sim
      \mathcal N\!\Bigl(
          e_k \mathrm{Down}(\boldsymbol{x}_1, 2^k),\; (1-e_k)^2 \boldsymbol{I}
      \Bigr),
  \\[2pt]
  \text{Start:}\quad
    \hat{\boldsymbol{x}}_{s_k} &\mid \boldsymbol{x}_1
      \sim
      \mathcal N\!\Bigl(
          s_k \mathrm{Up}\bigl(\mathrm{Down}(\boldsymbol{x}_1, 2^{k+1})\bigr),\;
          (1-s_k)^2 \boldsymbol{I}
      \Bigr).
\end{align*}
}

\noindent Importantly, a single model is used to learn the velocity field in each stage, allowing knowledge sharing between stages, unlike previous cascaded approaches which require separate models for generation and superresolution \citep{ho2022cascaded,ho2022imagen,wang2025lavie}.

\paragraph{Pyramidal Flow Matching} Pyramidal Flow Matching \citep{jin2024pyramidal} introduces pyramid flow matching in the video generation setting, sampling each frame with a pyramid flow autoregressively over time. This approach has important drawbacks which make it suboptimal for use in climate prediction, namely (i) the autoregressive framing leads to slower inference runtimes, and (ii) it depends on a pre-trained VAE to perform the flow efficiently in downsampled latent space. The approach we propose in this work can sample long temporal outputs much more efficiently without the need of a VAE by using parallel, funneled generation through multiple timescales.

\vspace{-1em}\paragraph{PixelFlow} More recently, PixelFlow \citep{chen2025pixelflow} adapts pyramid flows for image generation and removes the dependency on a VAE, so $\boldsymbol{x}$ lies in pixel space instead of latent space. Their approach demonstrates strong performance compared to slower, higher capacity models, but does not handle temporal data nor does the formulation handle heterogeneous resampling factors, which is necessary to model data at variable timescales as we further motivate below.

\subsection{Spatiotemporal Pyramid Flow}
\label{sec:spatiotemporal}
\begin{figure*}[t]
  \centering
  \includegraphics[width=0.92\textwidth]{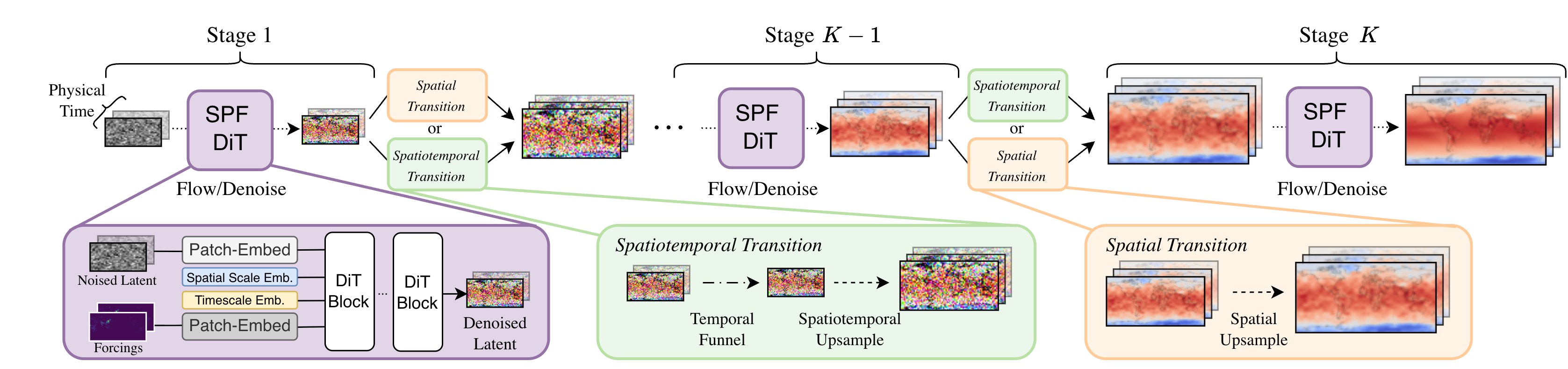}
  \caption{\textbf{SPF flow trajectory.} SPF divides generation into stages, each beginning with DiT denoising and followed by either a spatiotemporal transition (green) or a spatial-only transition (orange). Spatiotemporal transitions funnel into a timestep for the selected target period and upsample the latent in both space and time, while spatial transitions upsample only in space. This sequence of denoising and stage transitions continues until the final stage, which outputs clean samples at the target period and timescale.\vspace{-1em}}
  \label{fig:multiscale}
\end{figure*}

We introduce a class of pyramid flow models that cascade over both time and space within the flow, which we call spatiotemporal pyramid flows (SPFs). We first extend spatial pyramid flows, which previously only operate on spatial dimensions with homogeneous resampling factors,
to additionally handle and parallelize over a temporal dimension, as well as support heterogeneous resampling between stages (Figure~\ref{fig:overview}, \S~\ref{sec:temporal-flow}). To limit the high memory usage required to generate long sequences at the finest temporal resolution, we introduce a temporal funneling mechanism where we select a time period in the sequence and generate the next stage's noisy latents from the selected period
(\S~\ref{sec:funneled-flow}). Finally, to allow the approach to directly generate clean samples at any of the timescales in the pyramid, we adapt the piecewise flow matching procedure to include paths in which we only change the spatial resolution while keeping the temporal resolution fixed (\S~\ref{sec:multi-timescale}). Altogether, spatiotemporal pyramid flows enable temporally parallelized, direct sampling of sequences at multiple timescales, leading to efficient and accurate multiscale climate emulation (Figure~\ref{fig:multiscale}).

\subsubsection{Temporal Cascade and Heterogeneous Resampling}
\label{sec:temporal-flow}
To enable temporal generation, we generalize spatial pyramid flows to handle a temporal dimension. This requires changes to the probability paths and the procedure for handling jump points between stages during inference. To address this, we derive a general expression to handle renoising with a temporal dimension and any setting of resampling factors between stages.

\vspace{-1em}\paragraph{Conditional Probability Path}
We define a general form for the downsampling and upsampling layers for $k=0\,(finest)\dots K-1\,(coarsest)$ as:
\begin{align*}
  \mathrm{Down}_k(\boldsymbol{x}) &=
    \mathrm{Downsample}\bigl(\boldsymbol{x},\;\dot r_k^h,\;\dot r_k^w,\;\dot r_k^t\bigr), \\
  \mathrm{Up}_k(\boldsymbol{z}) &=
    \mathrm{Upsample}\bigl(\boldsymbol{z},\; r_{k+1}^h,\; r_{k+1}^w,\; r_{k+1}^t\bigr).
\end{align*}
where $r_k^h,r_k^w,r_k^t$ are the height, width, and temporal resampling factors between the $k$ and $k+t$th stage, and $\dot r_k^h=\prod_{i=1}^{k} r_i^h$, $\dot r_k^w=\prod_{i=1}^{k} r_i^w$, and
$\dot r_k^t=\prod_{i=1}^{k} r_i^t$ are cumulative factors.\\

\noindent Then the flow within the $k$th time window is:
\begin{equation*}
\boldsymbol{x}_t=t'\mathrm{Down}_k(\boldsymbol{x}_{e_k})+(1-t')\mathrm{Up}_k(\mathrm{Down}_{k+1}(\boldsymbol{x}_{s_k})),
\end{equation*}

\noindent with the conditional probability path defined by:
\begin{align}
  \text{End:}\quad
    \hat{\boldsymbol{x}}_{e_k} &\mid \boldsymbol{x}_1
      \sim
      \mathcal N\!\Bigl(
          e_k \mathrm{Down}_k(\boldsymbol{x}_1),\; (1-e_k)^2 \boldsymbol{I}
      \Bigr),
  \label{eq:end-dist-ours} \\[2pt]
  \text{Start:}\quad
    \hat{\boldsymbol{x}}_{s_k} &\mid \boldsymbol{x}_1
      \sim
      \mathcal N\!\Bigl(
          s_k \mathrm{Up}_k\bigl(\mathrm{Down}_{k+1}(\boldsymbol{x}_1)\bigr),\;
          (1-s_k)^2 \boldsymbol{I}
      \Bigr).
  \label{eq:start-dist-ours}
\end{align}

 \noindent Following \citet{jin2024pyramidal}, we couple the sampling of the probability path endpoints by enforcing the noise to be in the same direction. That is, we sample noise $n \sim \mathcal{N}(0,\boldsymbol{I})$ and then jointly compute the endpoints $(\hat{\boldsymbol{x}}_{e_k}, \hat{\boldsymbol{x}}_{s_k})$ with:
{\small
\begin{align}
\text{End:} \quad \hat{\boldsymbol{x}}_{e_k} &= e_k \mathrm{Down}_k(\boldsymbol{x}_1) + (1 - e_k) \boldsymbol{n}, \label{eq:end_sample} \\
\text{Start:} \quad \hat{\boldsymbol{x}}_{s_k} &= s_k \cdot \mathrm{Up}_k(\mathrm{Down}_{k+1}(\boldsymbol{x}_1)) + (1 - s_k) \cdot \boldsymbol{n}. \label{eq:start_sample}
\end{align}
}

\noindent Then we regress the flow model $v_t$ on the conditional vector field $u_t(\hat{\boldsymbol{x}}_t \mid \boldsymbol{x}_1) = \hat{\boldsymbol{x}}_{e_k} - \hat{\boldsymbol{x}}_{s_k}$ with the same flow matching objective, unifying generation and decompression:

\begin{equation}
\begin{aligned}
  \mathcal L_{\text{PFM}}
  &= \mathbb E_{\substack{
        k\sim \mathcal{U}\{0,\dots,K-1\},\\
        t\sim \mathcal{U}[s_k,e_k],\\
        (\hat{\boldsymbol{x}}_{e_k},\hat{\boldsymbol{x}}_{s_k})
     }}
     \biggl\|
        \boldsymbol{v}_t(\hat{\boldsymbol{x}}_t)
        - \bigl(\hat{\boldsymbol{x}}_{e_k}-\hat{\boldsymbol{x}}_{s_k}\bigr)
     \biggr\|^2 .
\end{aligned}
\label{eq:flow-matching-loss}
\end{equation}

\paragraph{Stage Continuity.}  
To ensure consistency at jump points between stages \cite{campbell2023trans}, we generalize the rescaling–renoising correction done in \citet{jin2024pyramidal} to handle arbitrary per-stage resampling factors and resampling along the temporal dimension. The resulting update rule is:
\begin{equation}
\hat{\boldsymbol{x}}_{s_k}
   = \frac{(1 - s_k) + s_k \sqrt{n_k}}{\sqrt{n_k}}
     \,\mathrm{Up}_k(\hat{\boldsymbol{x}}_{e_{k+1}})
     + (1 - s_k) \sqrt{\tfrac{n_k - 1}{n_k}}\, \boldsymbol{n}',
\label{eq:final-rescaling-renoising-main}
\end{equation}
where $n_k = r_k^h r_k^w r_k^t$ and $\boldsymbol{n}' \sim \mathcal{N}(0,I)$.  A full derivation is provided in the Appendix.

 Notably, Equation \eqref{eq:final-rescaling-renoising-main} ensures distributional continuity across stage transitions for \textit{any}
spatial and temporal resampling factors $r_k^h$, $r_k^w$, and $r_k^t$ (and in fact, for any number of dimensions). Hence the pyramid can follow natural timescale hierarchies of Earth system models, e.g.\ $\!{\times}10$ from decadal to yearly but $\!{\times}12$ from yearly
to monthly, while ensuring continuity at the jump points. In our implementation, we use $K=3$ stages to represent the commonly studied decadal, yearly, and monthly timescales, with $r_k^h=r_k^w=2$ and $r_1^t=10$, $r_2^t=12$. We note that the special cases of \citet{jin2024pyramidal} and \citet{chen2025pixelflow} use $r_k^h=r_k^w=2$ and $r_k^t=1$, i.e.\ purely spatial pyramids with homogeneous resampling factors of 2 per stage.

\subsubsection{Temporal Funneling}
\label{sec:funneled-flow}

As generating a long sequence of high spatial and temporal resolution outputs at once would be prohibitively expensive, and climate use cases often only require a subset of the full sequence for downstream analysis, we propose a method to \textit{funnel} into a time period of interest.

Specifically, we slice the latent before upsampling, saving memory and compute while still achieving probability path continuity. Because each denoised latent is Gaussian, selecting any subset of  \(T_k'\le T_k\) time indices from the denoised latent yields another Gaussian whose mean and covariance are simply the corresponding sub-vector and sub-matrix of the previous latent's parameters.
Therefore every rescaling–renoising step derived above applies unchanged and preserves the probability path with \(T_k'\) frames instead of \(T_k\). In our implementation, we funnel to a single frame, i.e.\ set \(T_k' = 1\). Since the temporal windows contain \(T_k\!\in\!\{10,12\}\) frames, this yields a 10 and 12$\times$ respective reduction in GPU memory and FLOPs per stage.

\subsubsection{Multi-Timescale Samples}
\label{sec:multi-timescale}

Climate use cases often require outputs at varying temporal resolutions (e.g. yearly, monthly, daily), but prior pyramid flows only allow for denoised samples at the highest resolution. To circumvent the slower procedure of generating at the highest temporal resolution then averaging, we adapt training and inference to allow for direct generation of clean samples at any timescale in the cascade.

\vspace{-1em}
\paragraph{Multi-Timescale Extension}
To allow the model to output clean samples directly at any timescale in the pyramid, we augment the flow matching objective
with an additional term evaluated on \emph{temporally frozen}
probability paths, where spatial resolution changes but temporal
resolution remains fixed. For each training instance, we draw
binary indicators $\delta_k \!\sim\! \mathrm{Bernoulli}(\pi_k)$ that
specify whether stage $k$ performs spatio-temporal refinement
($\delta_k \!=\! 1$) or spatial-only refinement ($\delta_k \!=\! 0$),
subject to the constraint that once a temporal transition occurs
($\delta_k \!=\! 1$), all finer stages remain temporal
($\delta_{k'} \!=\! 1$ for $k'\!>\!k$). For $K=3$ stages and denoting the $k$th timescale with $\mathcal{T}_k$, this produces three valid
temporal paths
($\mathcal{T}_0\!\rightarrow\mathcal{T}_0\!\rightarrow\mathcal{T}_0$,
$\mathcal{T}_0\!\rightarrow\mathcal{T}_1\!\rightarrow\mathcal{T}_1$, and
$\mathcal{T}_0\!\rightarrow\mathcal{T}_1\!\rightarrow\mathcal{T}_2$). To ensure balanced coverage
across these paths, we set the probabilities such that
$\delta_1 \!\sim\! \mathrm{Bernoulli}(2/3)$ and
$\delta_2 \mid (\delta_1\!=\!1) \!\sim\! \mathrm{Bernoulli}(1/2)$,
which yields equal probability for each path. In practice,
this entails randomly sampling minibatches that use different
flow matching pairs corresponding to the selected path.

\noindent
Given $\delta_k$, we define the per-stage temporal scaling factor
\[
\tilde r_k^{t} =
\begin{cases}
r_k^{t}, & \text{if } \delta_k = 1
\quad \text{(spatiotemporal refinement)}, \\[4pt]
1,       & \text{if } \delta_k = 0
\quad \text{(spatial-only refinement).}
\end{cases}
\]
Using
$\mathrm{Down}^{\star}_k =
\mathrm{Downsample}(\cdot;\dot r_k^{h},\dot r_k^{w},\tilde r_k^{t})$
and
$\mathrm{Up}^{\star}_k =
\mathrm{Up}(\cdot;r_{k+1}^{h},r_{k+1}^{w},\tilde r_{k+1}^{t})$
to denote the downsampling and upsampling layers under the sampled
path, the start–end pair for the conditional probability path is
\[
\begin{aligned}
  \hat{\boldsymbol{x}}_{s_k}^{\star} \;&=\;
  s_k\,\mathrm{Up}^{\star}_k\!\bigl(\mathrm{Down}^{\star}_{k+1}(\boldsymbol{x}_1)\bigr)
  + (1 - s_k)\boldsymbol{n}, \\
  \hat{\boldsymbol{x}}_{e_k}^{\star} \;&=\;
  e_k\,\mathrm{Down}^{\star}_k(\boldsymbol{x}_1) + (1 - e_k)\boldsymbol{n},
\end{aligned}
\]
reducing to
Eqs.~\eqref{eq:end_sample}–\eqref{eq:start_sample}
when selecting the path with spatial and temporal transitions occurring between every stage.

\vspace{-1em}
\paragraph{Multi-Timescale Objective}
With the definitions above,  the multi-timescale loss is
\begin{equation*}
  \mathcal L_{\mathrm{MT}}
  \;=\;
  \mathbb E_{k,t,\delta_k,(\boldsymbol{\hat{x}}^{\star}_{e_k},\boldsymbol{\hat{x}}^{\star}_{s_k})}
  \left\|
      \boldsymbol{v}_t(\hat{\boldsymbol{x}}_t)
      -\bigl(\hat{\boldsymbol{x}}^{\star}_{e_k}-\hat{\boldsymbol{x}}^{\star}_{s_k}\bigr)
  \right\|^{\!2}.
  \label{eq:mt-loss}
\end{equation*}
Because \(\delta_k\) is sampled on-the-fly, a single network is trained to denoise along any of the timescale paths.

\vspace{-1em}
\paragraph{Inference at Coarser Timescales}
At inference time, to directly sample coarser timescales (e.g. annual or decadal averages), we simply run the ODE solver up to the corresponding stage \(k^\star\),
apply the rescaling–renoising correction in Eq.~\eqref{eq:final-rescaling-renoising-main} to handle both spatial and temporal transitions, then only run spatial upsampling, adjusting the renoising correction accordingly, before finally denoising to generate the clean sample at the coarser timestep (Figure~\ref{fig:multiscale}). Importantly, this procedure enables generation of coarse samples without needing to first generate the finer temporal samples.

\paragraph{Spatial Scale and Timescale Embeddings} To allow the model to differentiate between stages at different spatial resolutions, we follow \citet{jin2024pyramidal} and use spatial scale embeddings with a fixed sinusoidal encoding passed through a 2-layer MLP with a SiLU activation \citep{elfwing2018sigmoid}. To allow the model to distinguish between inputs at different timescales with the same spatial resolution, we introduce a timescale embedding representing the current timescale of the flow. We use the same type of embedded sinusoidal encoding.

%% file: sec/3_data.tex
\begin{table*}[t]
\centering
\resizebox{0.98\linewidth}{!}{
\begin{tabular}{lccccccccccccc}
\toprule
\multirow{2}{*}{Model} & \lowerp{Temporal} & \lowerp{Native} & \multirow{2}{*}{Probabilistic} &
\multicolumn{4}{c}{\textit{Yearly}} & &
\multicolumn{4}{c}{\textit{Monthly}} \\
\cmidrule{5-8} \cmidrule{10-13}
 & \raisep{Outputs} & \raisep{Timescale} & & CRPS$\downarrow$ & RMSE$\downarrow$ & Bias$\lvert\downarrow\rvert$ & Runtime (s) && 
      CRPS$\downarrow$ & RMSE$\downarrow$ & Bias$\lvert\downarrow\rvert$ & Runtime (s) \\
\midrule

\multicolumn{11}{@{}l}{\textit{100M Parameters}} \\
\midrule
ClimaX Frozen \cite{nguyen2023climax} & Single & Yearly & \xmark & 0.405 & 0.619 & -0.014 & $<\!1$ && - & - & - & - \\
UNet \cite{ronneberger2015u}   & Single & Yearly & \xmark & 0.396 & 0.608 & -0.025 & $<\!1$ && - & - & - & - \\
ClimaX \cite{nguyen2023climax}  & Single & Yearly & \xmark & 0.347 & \textbf{0.546} & -0.073 & $<\!1$ && - & - & - & - \\
Pyramidal Flow \cite{jin2024pyramidal} & Autoregressive & Yearly & \cmark & 0.368 & 0.762 & -0.003 & 19 && - & - & - & -\\
Multi-Yearly Flow  (Ours)  & Parallel & Yearly & \cmark & 0.272 & 0.632 & 0.002 & 2 && - & - & - & - \\
Multi-Monthly Flow  (Ours)  & Parallel & Monthly & \cmark & 0.270 & 0.583 & -0.129 & 11 && \textbf{0.474} & \textbf{1.101} & -0.129 & 11\\
Pyramidal Flow \cite{jin2024pyramidal} & Autoregressive & Monthly & \cmark & 0.266 & 0.583 & -0.068 & 84 && 0.501 & 1.137 & -0.068 & 84\\
PixelFlow \cite{chen2025pixelflow}  & Single & Yearly & \cmark & \textbf{0.247} & \textbf{0.549} & -0.152 & 6 && - & - & - & - \\
\rowcolor{pastelgreen}
\lowerp{SPF (Ours)}  & \lowerp{Parallel}  & \lowerp{Multi} & \lowerp{\cmark} &
  \lowerp{\textbf{0.238}} & \lowerp{0.565} & \lowerp{0.004} & \lowerp{3} &&
  \lowerp{\textbf{0.462}} & \lowerp{\textbf{1.100}} & \lowerp{-0.043} & \lowerp{6}\\

\midrule
\multicolumn{11}{@{}l}{\textit{200M Parameters}} \\
\midrule
UNet \cite{ronneberger2015u}   & Single & Yearly & \xmark & 0.413 & 0.632 & -0.096 & 1 && - & - & - & - \\
Pyramidal Flow \cite{jin2024pyramidal} & Autoregressive & Yearly & \cmark & 0.327 & 0.671 & 0.035 & 24 && - & - & - & -\\
Multi-Yearly Flow (Ours) & Parallel & Yearly & \cmark & 0.254 & 0.595 & 0.002 & 4 && - & - & - & - \\
Pyramidal Flow \cite{jin2024pyramidal}  & Autoregressive & Monthly & \cmark & 0.236 & 0.532 & -0.034 & 105 && 0.473 & 1.083 & -0.034 & 105 \\
Multi-Monthly Flow (Ours)  & Parallel & Monthly & \cmark & 0.231 & 0.528 & 0.118 & 21 && \textbf{0.443} & \textbf{1.038} & 0.118 & 21 \\
PixelFlow \cite{chen2025pixelflow}  & Single & Yearly & \cmark & \textbf{0.224} & \textbf{0.504} & 0.033 & 7 && - & - & - & - \\
\rowcolor{pastelgreen}
\lowerp{SPF (Ours)}  & \lowerp{Parallel} & \lowerp{Multi} & \lowerp{\cmark} &
  \lowerp{\textbf{0.222}} & \lowerp{\textbf{0.511}} & \lowerp{-0.049} & 6 &&
  \lowerp{\textbf{0.453}} & \lowerp{\textbf{1.060}} & \lowerp{-0.063} & \lowerp{11}\\

\bottomrule
\end{tabular}}

\caption{\textbf{ClimateBench test metrics on held-out scenario (SSP2-4.5).}  Models use identical DiT backbones except for the UNet and ClimaX models, for which we match number of parameters.  We \textbf{bold} the top two results per parameter count and timescale. We report the runtime required to sample a single 10-year trajectory at each timescale on a single NVIDIA RTX 6000.}
\label{tab:climatebench}
\end{table*}

\section{Data}
\subsection{ClimateBench} ClimateBench \cite{watsonparris2022climatebench} is a benchmark dataset derived from NorESM2-LM, including historical simulations from 1850-2015 and future CMIP6 scenarios from 2015-2100. The outputs are spatially resolved annual anomalies of surface air temperature and precipitation on a 192$\times$288 latitude-longitude grid, with inputs consisting of four forcing variables on a lower resolution grid (96$\times$144): cumulative CO$_2$ and CH$_4$ emissions, and spatial maps of SO$_2$ and black carbon emissions. To match our expanded dataset ClimateSuite, we additionally include a stratospheric aerosol optical depth (AOD) input which uses historical AOD for historical scenarios and is set to 0 for all future scenarios. As ClimateBench was aggregated to yearly means, we acquire the monthly data (see Appendix) to evaluate models at a finer timescale. We train all models on the historical, SSP1-2.6, and SSP5-8.5 scenarios, use SSP3-7.0 for validation, and test on SSP2-4.5 as done in previous work \cite{nguyen2023climax,watsonparris2022climatebench}.

\subsection{ClimateSuite} 
Towards training better climate emulators, we introduce ClimateSuite, the largest climate-scale ML dataset to date containing 33,739 simulation-years of data from ten climate models. With data from multiple climate models, ClimateSuite facilitates the development of superemulators, or surrogates that can emulate multiple ESMs at once. Similar to ClimateBench, it includes historical simulations and future CMIP6 scenarios for each model. We include monthly resolution data for all scenarios in the dataset as is commonly recorded in climate model outputs. Appendix Table 5 compares ClimateSuite to existing climate-scale datasets.

ClimateSuite is the first ML dataset to include simulations of SAI, a proposed intervention to reduce the impacts of climate change by adding reflective particles to the upper atmosphere to reflect a small fraction of sunlight back to space. The SAI scenarios have varying date ranges, with most SAI simulated between 2035 and 2070. Appendix Table 6 enumerates all ClimateSuite scenarios and ESMs.

The inputs and outputs in ClimateSuite match those in ClimateBench. However, because the original four variables in ClimateBench do not capture the radiative signal from the stratospheric aerosols in SAI scenarios, we include an additional stratospheric aerosol optical depth (AOD) input, which quantifies the column-integrated extinction of sunlight by aerosols in the stratosphere. AOD thus serves as a direct proxy for the strength and spatial distribution of the forcing produced by SAI.

Because different climate models use different spatial resolutions, we interpolate all maps to the common 192 $\times$ 288 latitude-longitude grid used in CESM2-WACCM. We split into train, validation, and test following the scenario split used in our ClimateBench experiments, and hold out the UKESM1-0-LL SAI experiment to test generalization to a climate model for which SAI experiments were not in the training set. As ESMs have highly variable number of ensemble members, we select 3 members per climate model to balance among them. We provide more detail about how we curate and process ClimateSuite in the Appendix.

%% file: sec/4_experiments.tex
\section{Experiments}

\subsection{Experimental Details}
\subsubsection{Architecture}
We use the MM-DiT architecture from SD3 \cite{esser2024scaling} as the base model for all flow matching approaches. We consider two scales of the model, 100M parameters (2 layers) and 200M parameters (4 layers). We use sinusoidal position encodings \cite{vaswani2017attention} for the spatial dimensions following \cite{esser2024scaling} and 1D Rotary
Position Embeddings (RoPE) \cite{su2024roformer} 
for the temporal dimension following \cite{jin2024pyramidal}. For conditioning on the inputs, we use a learnable patch embedding, add the positional embeddings, then use standard cross-attention conditioning on the flattened patchified sequence each layer of the network. We condition each stage of the flow matching approaches with temporally aligned forcings from the corresponding timescale. We use a patch size of $8\times 8$ for the outputs and $16\times 16$ for the inputs as forcings exhibit smoother spatial structure and weaker local dependencies than the outputs. To allow for efficient attention across multiple spatial and temporal resolutions within a batch, we sequence pack after patchifying \cite{dehghani2023patch}, concatenating flattened token
sequences of variable lengths (due to different spatial and temporal resolutions) along the sequence dimension. These settings together allow us train and run inference on a single NVIDIA RTX A4000 with 16GB of VRAM. 

\subsubsection{Baselines}
We compare against several baselines at the yearly and monthly timescales. At the yearly timescale, we train UNets \cite{ronneberger2015u}, a fully fine-tuned ClimaX and ClimaX with a frozen backbone \cite{nguyen2023climax}, a PixelFlow model \cite{chen2025pixelflow}, a PyramidalFlow model \cite{jin2024pyramidal} which autoregressively predicts one year at a time over the coarse of one decade, and a single timescale version of our model that flows in parallel from noise at the yearly timescale to clean yearly samples spanning a decade, without any temporal funneling or resampling, which we refer to as Multi-Yearly Flow. The UNet, ClimaX, and PixelFlow models are non-temporal and output one year at a time independently. At the monthly timescale, we train a PyramidalFlow model \cite{jin2024pyramidal} which autoregressively predicts one month at a time over the coarse of one year. We also train a Multi-Monthly Flow which operates analogously to the Multi-Yearly Flow but instead generates monthly samples spanning a year. Therefore the Multi-Yearly and Multi-Monthly flow models output 10 and 12 frames respectively, matching SPF. We note all flow matching approaches operate in pixel space. To control for model capacity, we match the number of parameters of the UNet and use the same backbone architectures for the flow matching approaches. 

\subsubsection{Evaluation Procedure}
We evaluate with three standard metrics, namely root mean squared error (RMSE), bias, and continuous ranked probability score (CRPS). We weight all metrics by grid cell area, formally defined in the Appendix. RMSE and bias measure the quality of the mean prediction whereas CRPS captures the probabilistic skill of the model by assessing both the accuracy and sharpness of the predicted distribution relative to the targets. We evaluate the probabilistic models using 5 samples and standardize to 90 inference steps total, split equally by stage, using an Euler solver.

We evaluate the models at both monthly and yearly timescales. While we cannot evaluate the native yearly models at the monthly timescale, we additionally evaluate the native monthly models at the yearly timescale by averaging across the calendar year.

\subsection{Results}

\subsubsection{SPF Comparison to Baselines}
SPF achieves the lowest CRPS compared to all 100M and 200M parameter models respectively at the yearly timescale  (Table~\ref{tab:climatebench}). SPF with 200M params has the second lowest RMSE, close behind the yearly PixelFlow model and considerably ahead of all other 200M param models. With 100M params, it achieves slightly poorer RMSE than the single year PixelFlow and ClimaX models but outperforms all other approaches. We emphasize that ClimaX is a strong model pre-trained on a large climate dataset whereas SPF is only trained on the ClimateBench training set.  SPF's bias is close to perfect at the yearly timescale with 100M params, and comparable to other approaches with 200M params.  Importantly, it achieves this high performance with very efficient sampling: the 100M and 200M variants achieve $28\times$ and $>15\times$ speedups respectively compared to the autoregressive PyramidalFlow model, reducing sampling time down to a few seconds.

At the monthly timescale, SPF with 100M params achieves the best CRPS, RMSE, and bias, and the 200M param model achieves the lowest RMSE and close to best CRPS. Sampling 10 years of monthly outputs is $>3\times$ faster than PyramidalFlow with 200M params and $6\times$ faster with 100M. We plot SPF's monthly and yearly global means, histograms, and monthly and yearly samples compared to the simulation data in the Appendix.

\begin{table}[t]
\centering
\resizebox{\linewidth}{!}{%
\begin{tabular}{lccc}
\toprule
Variant & Stage Sequence  & CRPS$\downarrow$ & RMSE$\downarrow$ \\
\midrule
DYMMM       & Dec $\to$ Yr $\to$ Mo $\to$ Mo $\to$ Mo &  0.474 & 1.087         \\
DYYMM       & Dec $\to$ Yr $\to$ Yr $\to$ Mo $\to$ Mo & 0.463  & 1.085            \\
DYM-Monthly & Dec $\to$ Yr $\to$ Mo & 0.453 & 1.064\\
\rowcolor{pastelgreen} \lowerp{SPF (DYM-Any)}     & \lowerp{Dec $\to$ Yr $\to$ Mo}  & \lowerp{\textbf{0.453}} & \lowerp{\textbf{1.060}}\\
\bottomrule
\end{tabular}}
\caption{\textbf{Ablation study on the design of the spatiotemporal pyramid on ClimateBench SSP2-4.5 (monthly timescale)}.  SPF (DYM-Any) supports samples at any of the timescales, whereas the other variants only support monthly samples. All variants use an identical DiT architecture with 200M parameters.}
\label{tab:ablations}
\end{table}

\subsubsection{Pyramid Design Ablation}
We explore alternate pyramid designs of SPF. We consider an approach that uses the same three stage decadal $\rightarrow$ yearly $\rightarrow$ monthly design as SPF, but only trained to support clean monthly samples (DYM-Monthly). We also train two variants with decoupled spatial and temporal transitions, one starting with the temporal transitions and then ending with the spatial ones (DYMMM), and one where we alternate temporal and spatial transitions (DYYMM). We note that we considered an alternating pyramid that starts with spatial then temporal but this variant ran out of GPU memory.

SPF's design outperforms all other tested variants (Table~\ref{tab:ablations}). The 3 stage design without multi-scale sampling (DYM-Monthly) achieves similar CRPS but slightly higher RMSE than the SPF design (DYM-Any), indicating there is a small improvement to the monthly sampling performance when training the model to support sampling from the other timescales. The 5 stage variants underperform both 3 stage variants, with the alternating approach (DYYMM) outperforming the design starting with the temporal transitions (DYMMM). \\

\begin{table}[t]
\centering
\resizebox{0.975\linewidth}{!}{%
\begin{tabular}{ccccccc}
\toprule
\lowerp{ClimateSuite} & \multirow{2}{*}{Parameters} &
\multicolumn{2}{c}{\textit{Yearly}} & &
\multicolumn{2}{c}{\textit{Monthly}} \\
\cmidrule{3-4} \cmidrule{6-7}
\raisep{Pretrained} & & CRPS & RMSE && CRPS & RMSE \\
\midrule
- &  200M & 0.222 & 0.511 &&  0.453 & 1.060 \\
- &  600M & 0.229 & 0.523 && 0.442 &  1.053\\
\checkmark &  600M & \textbf{0.216} & \textbf{0.491} &&  \textbf{0.432 }&  \textbf{1.026}\\
\bottomrule
\end{tabular}}
\caption{\textbf{Effect of scale and ClimateSuite pre-training on ClimateBench performance.}}
\label{tab:scale}
\end{table}

\vspace{-1em}
\subsubsection{Scaled SPF Multi-Model Results}
To investigate the impact of scale and ClimateSuite pre-training, we train a 600M parameter SPF model on ClimateSuite then fine-tune it on ClimateBench. We find that scale and pre-training together lead to improvements across the yearly and monthly timescales (Table~\ref{tab:scale}). This gain is not only due to scale, as a 600M parameter non-pre-trained model underperforms the 200M parameter variant at the yearly timescale, and the pre-trained model outperforms the non-pre-trained model on both timescales. 

We additionally evaluate SPF against a 600M-parameter UNet explored previously for multi-model emulation \cite{kaltenborn2023climateset} on the held-out SSP scenario across 10 climate models and the held-out SAI scenario on UKESM1-0-LL. SPF achieves superior performance across all climate models (Table \ref{tab:climateset_combined}). Notably, on the held-out SAI scenario, SPF generates samples with global means close to the simulations in most of the century, with poorer performance at the start. However, we note that this is achieved without training on any SAI data from UKESM1-0-LL.  We plot global means, histograms, and samples per model compared to the ground truth simulations in the Appendix.

\begin{table}[t]
\centering
\resizebox{0.9\linewidth}{!}{%
\begin{tabular}{lcccc}
\toprule
\multirow{2}{*}{Climate Model} &
\multicolumn{2}{c}{RMSE $\downarrow$} &
\multicolumn{2}{c}{CRPS $\downarrow$} \\
\cmidrule(lr){2-3}\cmidrule(lr){4-5}
& UNet & SPF & UNet & SPF \\
\midrule
\multicolumn{5}{l}{\textit{Standard scenario (SSP2-4.5)}} \\
\midrule 
\addlinespace[2pt]
BCC-CSM2-MR     & 0.770 & \textbf{0.749} & 0.491 & \textbf{0.311} \\
CESM2           & 0.523 & \textbf{0.473} & 0.326 & \textbf{0.224} \\
CESM2-WACCM     & 0.521 & \textbf{0.468} & 0.332 & \textbf{0.216} \\
CMCC-CM2-SR5    & 0.842 & \textbf{0.745} & 0.539 & \textbf{0.318} \\
CMCC-ESM2       & 0.886 & \textbf{0.817} & 0.552 & \textbf{0.347} \\
GFDL-ESM4       & 0.520 & \textbf{0.483} & 0.320 & \textbf{0.223} \\
IPSL-CM6A-LR    & 0.544 & \textbf{0.432} & 0.330 & \textbf{0.200} \\
MRI-ESM2-0      & 0.540 & \textbf{0.502} & 0.347 & \textbf{0.230} \\
NorESM2-LM      & 0.536 & \textbf{0.491} & 0.350 & \textbf{0.226} \\
UKESM1-0-LL     & 0.599 & \textbf{0.561} & 0.343 & \textbf{0.261} \\
\cmidrule(lr{0.3em}){2-5}
Average          & 0.616 & \textbf{0.573} & 0.393 & \textbf{0.256} \\
\midrule
\multicolumn{5}{l}{\textit{Intervention scenario (SAI)}} \\
\midrule 
\addlinespace[2pt]
UKESM1-0-LL     & 0.883 & \textbf{0.727} & 0.466 & \textbf{0.315} \\
\bottomrule
\end{tabular}}
\caption{\textbf{Yearly metrics on held-out scenarios across climate models in ClimateSuite.}
CRPS is evaluated using 5 ensemble members. Both UNet and SPF use 600M parameters. We \textbf{bold} the better value per row.}
\label{tab:climateset_combined}
\end{table}

%% file: sec/5_discussion.tex
\section{Conclusion}
We introduce SPF, a new flow matching approach to model spatiotemporal data at multiple timescales. Our results show that SPF is an efficient and accurate probabilistic climate emulator, enabling scalable and high-fidelity modeling of multiple ESMs, future scenarios, and timescales. We also curate ClimateSuite, the largest ML-ready climate-scale dataset. We hope that ClimateSuite will serve as a valuable resource to train multi-model superemulators and help mitigate the overfitting observed in prior models trained on the much smaller ClimateBench dataset \cite{lutjens2025impact}. Together, SPF and ClimateSuite advance the development of scalable, probabilistic emulators for ESMs.

Our work has limitations. First, SPF does not explicitly enforce physical constraints such as energy or mass conservation, which may yield physically inconsistent trajectories under some forcing conditions. Second, while ClimateSuite spans multiple models and scenarios, it is constrained to existing ESM ensembles, likely limiting generalization to unseen parameterizations or extreme forcings. Future work should explore ways to overcome these limitations.

%% file: sec/6_suppl.tex
\clearpage
\appendix 
\section*{Appendix Outline}
\addcontentsline{toc}{section}{Appendix Outline}
\startcontents[appendix]
\printcontents[appendix]{l}{1}{}

\renewcommand\thefigure{S\arabic{figure}}    
\renewcommand\thetable{S\arabic{table}}
\setcounter{figure}{0}
\setcounter{table}{0}

\section{Related Work}

\subsection{Diffusion and Flow Matching Models}
\noindent Diffusion models have become a cornerstone of generative modeling, framing synthesis as the reversal of a gradual noising process through learned score functions \cite{ho2020denoising,song2019generative}. Subsequent work extended this paradigm to conditional and guided settings, achieving state-of-the-art image generation \cite{dhariwal2021diffusion} and later to spatiotemporal domains via video diffusion and latent video variants for efficient long-horizon synthesis \cite{ho2022video,zhou2022magicvideo}. To further scale quality and resolution, cascaded diffusion pipelines decompose generation into multi-stage refinement processes, where coarse outputs are progressively upsampled by higher-resolution diffusion models \cite{ho2022cascaded,saharia2022image}. Flow-based models offer an alternative by learning invertible mappings from noise to data with exact likelihoods \cite{dinh2016density,kingma2018glow}, and the introduction of flow matching unified flow and diffusion training through deterministic ODE trajectories between distributions \cite{lipman2022flow}. Recent hierarchical flow architectures extend this principle across scales: PyramidalFlow \cite{jin2024pyramidal} and PixelFlow \cite{chen2025pixelflow} use pyramids of flows at increasing spatial or spatiotemporal resolutions to more efficiently model video and image data respectively.

\subsection{Generative Models for Climate}
Generative AI has increasingly been applied in climate science to emulate expensive simulations and enhance spatial resolution. Diffusion models, in particular, have shown strong potential for generating realistic spatiotemporal climate fields. Several previous works have demonstrated that diffusion models can be used for downscaling, to enhance coarse reanalysis and model fields to higher resolutions \citet{watt2024generative,ling2024diffusion,srivastava2024precip}. DiffESM applies conditional diffusion to produce daily climate variables consistent with coarse monthly means \cite{bassetti2024diffesm}. Climate in a Bottle (cBottle) proposes a two-stage diffusion framework that first synthesizes coarse 100~km atmospheric fields before applying a learned diffusion-based super-resolution to reach kilometer scales \cite{brenowitz2025climate}. Spherical DYffusion introduces a weather-scale probabilistic emulator based on a dynamics-informed `dyffusion' process paired with a spherical Fourier neural operator, enabling stable long-horizon global climate emulation of a simplified atmospheric model with fixed forcings \cite{cachay2024sphericaldyffusion}. 

\section{Rescaling and Renoising for Probability Path Continuity}
\label{app:spatiotemporal-derivation}
We derive the rescaling-renoising correction described in Equation 6 to handle the additional temporal dimension and support hetereogeneous resampling factors between stages.

\subsection{Spatial-Only Derivation}
We will start with the simpler spatial-only, homogeneous resampling setting presented in Section 2.1.3 and rederive the intermediate steps from \citet{jin2024pyramidal} here for clarity. Following \citet{jin2024pyramidal}, we upsample the previous low-resolution endpoint using nearest-neighbor resampling, resulting in a linear combination of the inputs, which therefore follows a Gaussian distribution:
\begin{equation}
\resizebox{0.95\columnwidth}{!}{$
\mathrm{Up}_k(\hat{\boldsymbol{x}}^{e_{k+1}}) \mid \boldsymbol{x}_1 \sim \mathcal{N}\left(e_{k+1} \mathrm{Up}_k(\mathrm{Down}_{k+1}(\boldsymbol{x}_1)), (1 - e_{k+1})^2 \boldsymbol{\Sigma} \right),
$}
\label{eq:upsample-gaussian}
\end{equation}
with $\boldsymbol{\Sigma}$ is the covariance matrix induced by the upsampling operation.\\

\noindent To ensure continuity of the probability path between different stages of the spatial pyramid, the endpoints must have the same distributions. Eqs. (1), (2), and \eqref{eq:upsample-gaussian} show that the distributions of the endpoints are similar after a simple upsampling transformation:
{\small
\begin{align}
\hat{\boldsymbol{x}}_{s_k} \mid \boldsymbol{x}_1 &\sim \mathcal{N}\left(s_k  \mathrm{Up}_k(\mathrm{Down}_{k+1}(\boldsymbol{x}_1)),\, (1 - s_k)^2 I\right) \label{eq:start_endpoint}\\
\mathrm{Up}_k(\hat{\boldsymbol{x}}_{e_{k+1}}) \mid \boldsymbol{x}_1 &\sim \mathcal{N}\left(e_{k+1} \mathrm{Up}_k(\mathrm{Down}_{k+1}(\boldsymbol{x}_1)), (1 - e_{k+1})^2 \boldsymbol{\Sigma} \right).
\end{align}
}

\noindent We can therefore apply a linear transformation with a corrective Gaussian noise to match the distributions:
\begin{equation}
\hat{\boldsymbol{x}}_{s_k} = \frac{s_k}{e_{k+1}} \text{Up}_k(\hat{\boldsymbol{x}}_{e_{k+1}}) + \alpha \boldsymbol{n}', \quad \text{where } \boldsymbol{n}' \sim \mathcal{N}(0, \boldsymbol{\Sigma}').
\label{eq:linear_correction}
\end{equation}

\noindent The rescaling coefficient $\frac{s_k}{e_{k+1}}$ matches the means of these distributions, and $\alpha$ is the noise coefficient. To determine $\alpha$ and $\boldsymbol{\Sigma}'$, we need to match the covariance matrices of Eqs.~\eqref{eq:start_endpoint} and \eqref{eq:linear_correction}
\begin{equation}
\frac{s_k^2}{e_{k+1}^2} (1 - e_{k+1})^2 \boldsymbol{\Sigma} + \alpha^2 \boldsymbol{\Sigma}' = (1 - s_k)^2 \boldsymbol{I}.
\label{eq:covariance_matching}
\end{equation}

\noindent \citet{jin2024pyramidal} show that for a simple nearest neighbor upsampling operation, $\boldsymbol{\Sigma}$ and $\boldsymbol{\Sigma}'$ have blockwise structure that can be exploited to determine the noise correction to match the endpoint distributions.

\begin{table*}[t]
\centering
\resizebox{\linewidth}{!}{
\begin{tabular}{lccccccc}
\toprule
\multirow{2}{*}{Dataset} & Climate & Model & Model $\times$ Scenarios  & Intervention Scenarios & Simulation & \multicolumn{2}{c}{Resolution}\\
&Models & $\times$ Scenarios & $\times$ Members & $\times$ Members & Years & Spatial & Temporal\\
\midrule
ClimateBench \cite{watsonparris2022climatebench} & 1 & 5 & 11 & 0 & 1,183 & 192 $\times$ 288 & Monthly\\
ACE \cite{wattmeyer2024ace} & 1 & 1 & 1 & 0 & 110 & 180 $\times$ 360 & 6-Hourly\\
ClimateSet \cite{kaltenborn2023climateset} & 21 & 104 & 108 & 0 & 10,672 & 192 $\times$ 288 & Daily\\
ERA5 \cite{hersbach2020era5} & 1 & 1 & 1 & 0 & 85 & 720 $\times$ 1440 & Hourly\\
\midrule 
ClimateSuite (Ours) & 10 & 66 & 345 & 69  & 33,739 & 192 $\times$ 288 & Monthly\\
\bottomrule
\end{tabular}
}
\caption{\textbf{Comparison of ClimateSuite to existing climate-scale datasets.}}
\label{tab:datasets}
\end{table*}

\begin{table}[t!]
    \centering
    \begin{tabular}{lccc}
    \toprule
    Climate Model & Historical / SSP & SAI & Years \\
    \midrule
    BCC-CSM2-MR     & 5 & 0  & 839  \\
    CESM2           & 5 & 0  & 2847 \\
    CESM2-WACCM     & 5 & 15 & 3522 \\
    CMCC-CM2-SR5    & 5 & 0  & 2159 \\
    CMCC-ESM2       & 5 & 0  & 509  \\
    GFDL-ESM4       & 5 & 0  & 1011 \\
    IPSL-CM6A-LR    & 5 & 0  & 8455 \\
    MRI-ESM2-0      & 5 & 0  & 4216 \\
    NorESM2-LM      & 5 & 0  & 2294 \\
    UKESM1-0-LL     & 5 & 1  & 7887 \\
    \midrule
    \textbf{Total}  & \textbf{50} & \textbf{16} & \textbf{33,739} \\
    \bottomrule
    \end{tabular}
    \caption{\textbf{Climate model and scenario breakdown in ClimateSuite.}}
    \label{tab:climatesuite}
\end{table}

\subsection{Temporal and Hetegeneous Resampling Derivation}

We now consider the setting presented in Section 2.2.1. We generalize these covariance matrices to any differing, per-stage resampling factors $r_k^h$, $r_k^w$, and $r_k^t$ by allowing per-stage upsampling covariance matrices $\boldsymbol{\Sigma}_k$ and corrective noise covariance matrices $\boldsymbol{\Sigma}_k'$ as well as per-stage corrective noise coefficients $\alpha_k$. Rewriting Eqs~\eqref{eq:linear_correction} and \eqref{eq:covariance_matching},

\begin{equation}
\hat{\boldsymbol{x}}_{s_k} = \frac{s_k}{e_{k+1}} \mathrm{Up}_k(\hat{\boldsymbol{x}}_{e_{k+1}}) + \alpha_k \boldsymbol{n}', \quad \text{where } \boldsymbol{n}' \sim \mathcal{N}(0, \boldsymbol{\Sigma}_k').
\label{eq:linear_correction-general}
\end{equation}
\begin{equation}
\frac{s_k^2}{e_{k+1}^2} (1 - e_{k+1})^2 \boldsymbol{\Sigma}_k + \alpha_k^2 \boldsymbol{\Sigma}_k' = (1 - s_k)^2 \boldsymbol{I}.
\label{eq:covariance_matching-general}
\end{equation}

\noindent Importantly, the per-stage covariance matrices remain $n_k\times n_k$ block diagonal with $n_k=r_k^h\cdot r_k^w\cdot r_k^t$. 

\begin{align}
(\boldsymbol{\Sigma}_k)_{\text{block}}
  &= \mathbf{J}_n =
    \begin{pmatrix}
      1 & 1 & \dots & 1\\
      1 & \ddots & \ddots & \vdots \\
      \vdots & \ddots & \ddots & 1 \\
      1 & \dots & 1 & 1
    \end{pmatrix}, \notag \\[6pt]
\Rightarrow\quad
(\boldsymbol{\Sigma}'_k)_{\text{block}}
  &= \begin{pmatrix}
      1 & \gamma_k & \dots & \gamma_k\\
      \gamma_k & \ddots & \ddots & \vdots \\
      \vdots & \ddots & \ddots & \gamma_k \\
      \gamma_k & \dots & \gamma_k & 1
    \end{pmatrix}.
\label{eq:covariance-structure-general}
\end{align}

\noindent where $\gamma_k$ is a negative value in $[-\tfrac{1}{n_k-1}, 0]$ for decorrelation, and its lower bound ensures that the covariance matrix is positive semidefinite i.e. forms an equicorrelation matrix.\\

\noindent We can then rewrite Eqs.~\eqref{eq:covariance_matching-general} and \eqref{eq:covariance-structure-general} by equating their diagonal and non-diagonal elements respectively:
\begin{align*}
\frac{s_k^2}{e_{k+1}^2} (1 - e_{k+1})^2 + \alpha_k^2 &= (1 - s_k)^2, \\
\frac{s_k^2}{e_{k+1}^2} (1 - e_{k+1})^2 + \alpha_k^2 \gamma_k &= 0.
\end{align*}

\noindent Since $0 < s_k, e_{k+1} < 1$, the equations are solvable with:
\begin{equation}
e_{k+1} = \frac{s_k \sqrt{1 - \gamma_k}}{(1 - s_k)\sqrt{-\gamma_k} + s_k \sqrt{1 - \gamma_k}}, \quad
\alpha_k = \frac{1 - s_k}{\sqrt{1 - \gamma_k}}.
\label{eq:e_alpha_solved}
\end{equation}

\noindent Intuitively, we want to preserve signals maximally at each jump point, which corresponds to minimizing the noise weight $\alpha_k$. According to Eq.~\eqref{eq:e_alpha_solved}, this is equivalent to minimizing $\gamma_k$ at each jump point. Substituting the minimum value $\gamma_k = -\tfrac{1}{n_k-1}$ into Eq.~\eqref{eq:e_alpha_solved} yields:

\begin{equation*}
e_{k+1} = \frac{s_k \sqrt{n_k}}{(1 - s_k) + s_k \sqrt{n_k}}, \quad
\alpha_k = (1 - s_k) \sqrt{\frac{n_k - 1}{n_k}}
\end{equation*}

\noindent As in \citet{jin2024pyramidal}, we find that $e_{k+1} > s_k$ in this generalized form, meaning the timestep is rolled back when adding the corrective noise at each jump point. Substituting this back into equation~\eqref{eq:linear_correction-general} yields
{\small
\begin{align*}
\hat{\boldsymbol{x}}_{s_k} &= \frac{s_k}{e_{k+1}} \mathrm{Up}_k(\hat{\boldsymbol{x}}_{e_{k+1}}) + \alpha_k \boldsymbol{n}'\\
&= \frac{(1 - s_k) + s_k \sqrt{n_k}}{\sqrt{n_k}} \mathrm{Up}_k(\hat{\boldsymbol{x}}_{e_{k+1}}) + (1 - s_k) \sqrt{\frac{n_k - 1}{n_k}} \boldsymbol{n}'
\end{align*}
}
matching the update rule presented in Eq. (6).

\begin{table*}
\resizebox{0.95\linewidth}{!}{
\begin{tabular}{lp{10cm}}
\toprule
Scenario & Description \\
\midrule
historical & Standard historical simulation from 1850 to near-present using observed forcings \\
ssp126 & Future scenario under SSP1-2.6, representing strong mitigation and low greenhouse-gas forcing \\
ssp245 & Future scenario under SSP2-4.5, a middle-of-the-road pathway of moderate emissions \\
ssp370 & Future scenario under SSP3-7.0 representing regional rivalry and higher forcing \\
ssp585 & Future scenario under SSP5-8.5, representing high emissions and fossil-fuel driven development \\
MA-HISTORICAL & Historical baseline simulation (pre-intervention control run) \\
MA-BASELINE & Baseline simulation without any intervention or injection applied, under future forcing \\
SINGLE-POINT-INJANN0N\_12Tg & Single-point injection at 0$^\circ$N (equator), 12 Tg total \\
SINGLE-POINT-INJANN15S\_12Tg & Single-point injection at 15$^\circ$S, 12 Tg total \\
SINGLE-POINT-INJANN15N\_12Tg & Single-point injection at 15$^\circ$N, 12 Tg total \\
SINGLE-POINT-INJANN30N\_12Tg & Single-point injection at 30$^\circ$N, 12 Tg total \\
SINGLE-POINT-INJANN30S\_12Tg & Single-point injection at 30$^\circ$S, 12 Tg total \\
SINGLE-POINT-INJMAM60N\_12Tg & Single-point injection at 60$^\circ$N, 12 Tg total \\
SINGLE-POINT-INJSON60S\_12Tg & Single-point injection at 60$^\circ$S, 12 Tg total \\
SSP245-MA-GAUSS-DEFAULT & Injection with a controller-based algorithm at 15$^\circ$S, 15$^\circ$N, 30$^\circ$S, 30$^\circ$N to maintain temperature near 1.5$^\circ$C above pre-industrial (PI) levels \\
SSP245-MA-GAUSS-LOWER-0.5 & Injection with a controller-based algorithm targeting 0.5$^\circ$C above PI\\
SSP245-MA-GAUSS-LOWER-1.0 & Injection with a controller-based algorithm targeting 1.0$^\circ$C above PI\\
SSP245-MA-GAUSS15N\_15S-LOWER-0.5 & Two-point injection at $\pm$15$^\circ$N/S targeting 0.5$^\circ$C above PI\\
SSP245-MA-GAUSS30N\_30S-LOWER-0.5 & Two-point injection at $\pm$30$^\circ$N/S targeting 0.5$^\circ$C above PI\\
SSP245-MA-GAUSS0N-LOWER-0.5 & Single-point injection at 0$^\circ$N targeting 0.5$^\circ$C above PI\\
SAI-1.5 & Multi-latitude injection targeting 1.5$^\circ$C above PI\\ 
\bottomrule
\end{tabular}
}
\caption{\textbf{Standard emissions and intervention scenarios included in ClimateSuite.} Both historical simulations use CESM2-WACCM and all non-historical SAI simulations use CESM2-WACCM and SSP2-4.5 as the base forcing scenario, except for SAI-1.5 which uses UKESM1-0-LL (under SSP2-4.5).}
\label{tab:climatesuite-sai}
\end{table*}

\section{ClimateSuite Dataset}
\label{app:climatesuite}

\subsection{Climate Model and Scenario Selection}
We select ten CMIP6 Earth system models to ensure broad coverage across model families, physical parameterizations, and scenario availability (Table~\ref{tab:climatesuite}). We include NorESM2-LM specifically because it is the reference model used in ClimateBench, enabling direct comparison with prior benchmarks. CESM2-WACCM and UKESM1-0-LL are two of the only ESMs for which fully coupled SAI intervention experiments have been conducted. The remaining models, namely BCC-CSM2-MR, CESM2, CMCC-CM2-SR5, CMCC-ESM2, GFDL-ESM4, IPSL-CM6A-LR, and MRI-ESM2-0, are widely used, well-validated CMIP6 models that together provide a diverse set of dynamical cores and physical parameterizations, increasing robustness to structural uncertainty. Importantly, each of these models provides simulations for historical and standard SSP scenarios (1-2.6/2-4.5/3-7.0/5-8.5) as well as multiple ensemble members. This collection therefore spans a representative cross-section of CMIP6 modeling centers while supporting both standard climate scenarios and specialized SAI experiments.

We ensure broad coverage of standard climate trajectories and intervention scenarios (Table~\ref{tab:climatesuite-sai}). We include historical simulations and four SSP scenarios spanning strong mitigation (SSP1-2.6), moderate emissions (SSP2-4.5), regional-rivalry–driven warming (SSP3-7.0), and high-emissions futures (SSP5-8.5). These scenarios collectively capture a range of plausible 21st-century forcing pathways used in CMIP6 and provide diverse data for training and evaluating models under a variety radiative environments. To study climate responses under SAI, we further incorporate historical and baseline control runs in addition to a variety of single-point, two-point, and multi-latitude injection strategies. These experiments span equatorial to high-latitude injections, fixed-mass versus controller-based deployment algorithms, and temperature targets ranging from 0.5°C to 1.5°C above pre-industrial levels. By grounding all intervention runs in a common SSP2-4.5 forcing scenario, we isolate the effect of aerosol injection while maintaining comparability across experimental designs. Together, this collection is a representative and scientifically comprehensive set of forcings that enables evaluation of models across conventional, extreme, and policy-relevant climate futures.

\subsection{Simulation Downloading}
We acquire all datasets using publicly accessible portals and tools. We download external forcings from input4MIPs through the ESGF portal, ensuring consistency with the forcing datasets used in the original modeling center runs. For the standard CMIP6 historical and SSP simulations, we use ESMValCore \cite{Andela_ESMValCore_2025} and acccmip6 \cite{Hassan_acccmip6_Python_package_2022} to automate search, download, and integrity checks across ten models. We use both tools to increase coverage because each exposes a partially overlapping subset of CMIP6 replica servers. For the SAI experiments, we retrieve simulations via Globus. We obtain the UKESM1-0-LL SAI outputs directly from the Met Office ARISE portal \cite{MetOffice_arise-cmor-tables_2025}.

\subsection{Data Processing}
We process all data to standardized NetCDFs to provide an ML-friendly format while still facilitating common climate analysis. To standardize the spatial resolution among all climate models, we regrid with bilinear interpolation and enforce longitudinal periodicity, preserving large-scale spatial patterns to ensure physically smooth transitions between adjacent grid cells.

\section{Evaluation Metrics}
\label{app:eval-metrics}
\subsection{Preliminaries}
Let $\mathbf{X} \in \mathbb{R}^{E \times I \times J}$ denote an ensemble of predictions (i.e. multiple samples from different starting noise, all using the same conditioning), and $\mathbf{Y} \in \mathbb{R}^{I \times J}$ the corresponding simulation targets, where $E$ is the number of ensemble members, $I$ the number of latitudes, and $J$ the number of longitudes in the grid. Let $\mathbf{\overline{X}}=\frac1E\sum_{e=1}^E$ be the prediction averaged over ensemble members. Define $w(i)$ to be the normalized latitude-dependent area weight at latitude $i$, such that
\[
\frac{1}{I}\sum_{i=1}^{I} w(i) = 1,
\]
These weights account for the decreasing surface area of grid cells toward the poles. They are therefore used to compute spatially unbiased means and evaluation metrics, a standard practice in weather and climate prediction \cite{nguyen2023climax,watsonparris2022climatebench,cachay2024sphericaldyffusion,rasp2020weatherbench}.

\subsection{Metric definitions}
We report the area-weighted average Root Mean Square Error (RMSE) and Bias of the member-averaged predictions compared to the targets as follows:
\begin{align}
\text{Bias} &= \frac{1}{IJ} \sum_{i,j} w(i)\big(\mathbf{\overline{X}}_{i,j} - \textbf{Y}_{i,j}\big), \label{eq:bias} \\[4pt]
\text{RMSE} &= \sqrt{\frac{1}{IJ}\sum_{i,j} w(i)\big(\mathbf{\overline{X}}_{i,j} - \textbf{Y}_{i,j}\big)^2}. \label{eq:rmse}
\end{align}
Values closer to zero are better for Bias and lower values are better for RMSE.\\


\noindent We additionally evaluate the full ensemble forecast using the unbiased version of the CRPS~\cite{matheson1976scoring}:
\begin{equation}
\begin{split}
\text{CRPS} = \frac{1}{IJ}\sum_{i,j} w(i) \Bigg[
\frac{1}{E}\sum_{e=1}^{E} \big|\mathbf{X}_{e,i,j} - \mathbf{Y}_{i,j}\big| \\
- \frac{1}{2E(E-1)}\sum_{e=1}^{E}\sum_{f=1}^{E}
\big|\mathbf{X}_{e,i,j} - \mathbf{X}_{f,i,j}\big|
\Bigg].
\label{eq:crps}
\end{split}
\end{equation}

\begin{figure*}[t!]
    \centering
    \includegraphics[width=0.95\linewidth]{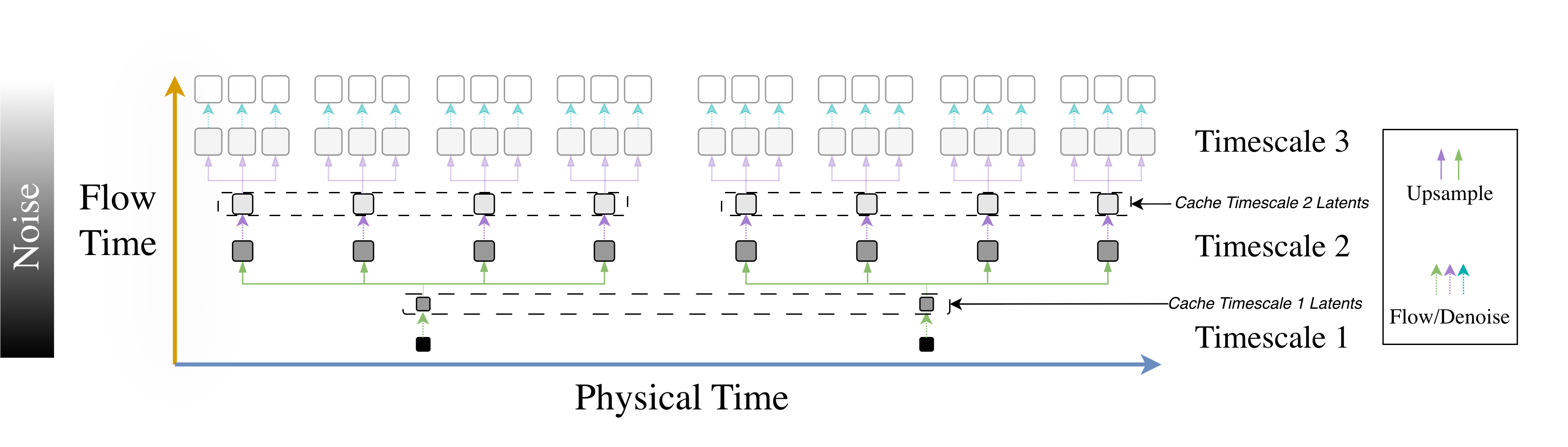}
    \caption{\textbf{Efficiency benefits from caching.} For long sequence generation, intermediate latents at coarser timescales can be cached to save compute when generating samples from finer timescales. In this dummy example, to generate a sequence of eight fine timescale samples, the flow in Timescale 1 only needs to be run once and the flow in Timescale 2 only needs to be run twice. Then, only the last part of the flow in Timescale 3 needs to be run when generating each of the eight clean Timescale 3 samples as the flow can resume from the Timescale 2 latents.}
    \label{fig:caching}
\end{figure*}

\noindent The first term represents the accuracy (or skill) of the ensemble, while the second term quantifies its internal spread.  
The conventional, biased form of CRPS averages the spread using a factor of $\frac{1}{2E^2}$, which introduces bias for small ensembles.  
The unbiased formulation instead employs $\frac{1}{2E(E-1)}$, providing a more reliable measure of ensemble performance.  
Lower CRPS values indicate better forecasts.  
For deterministic models ($E=1$), $\mathbf{\overline{X}}_{i,j}=\mathbf{X}_{i,j}$ and CRPS simplifies to the mean absolute error (MAE).

\begin{table}[t]
\centering
\resizebox{0.98\linewidth}{!}{
\begin{tabular}{lccc}
\toprule
Model & Native Timescale & Yearly Runtime (s) & Monthly Runtime (s)\\
\midrule

\multicolumn{3}{@{}l}{\textit{100M Parameters}} \\
\midrule
ClimaX Frozen \cite{nguyen2023climax} & Yearly & 2 & - \\
UNet \cite{ronneberger2015u} & Yearly & 1 & -\\
ClimaX \cite{nguyen2023climax} & Yearly & 2 & -\\
Pyramidal Flow \cite{jin2024pyramidal}& Yearly & 190 & - \\
Multi-Yearly Flow  (Ours) & Yearly & 20 & -\\
Multi-Monthly Flow  (Ours) & Monthly & 112 & 112\\
Pyramidal Flow \cite{jin2024pyramidal} & Monthly & 844 & 844\\
PixelFlow \cite{chen2025pixelflow} & Yearly & 60 & -\\
\rowcolor{pastelgreen}
\lowerp{SPF (Ours)} & Multi & \lowerp{21} & \lowerp{52}\\

\midrule
\multicolumn{3}{@{}l}{\textit{200M Parameters}} \\
\midrule
UNet \cite{ronneberger2015u} & Yearly & 9 & - \\
Pyramidal Flow \cite{jin2024pyramidal} & Yearly & 239 & -\\
Multi-Yearly Flow (Ours) & Yearly & 41 & -\\
Pyramidal Flow \cite{jin2024pyramidal} & Monthly & 1054 & 1054\\
Multi-Monthly Flow (Ours) & Monthly & 213  & 213\\
PixelFlow \cite{chen2025pixelflow} & Yearly & 71 & -\\
\rowcolor{pastelgreen}
\lowerp{SPF (Ours)} & Multi & \lowerp{42} & \lowerp{100}\\

\bottomrule
\end{tabular}}
\caption{\textbf{Per-timescale runtime for a single 100 year sample.}}
\label{tab:100year-runtimes}
\end{table}

\section{Additional Results}
\label{sec:additional-results}

\subsection{Long Sequence Runtimes}
A major benefit of SPF's design is the capability of caching intermediate states to save computation for long sequences. We show this visually for a dummy example in Figure~\ref{fig:caching}. To demonstrate this empirically, we report the runtimes for generating a 100-year sequence at both the yearly and monthly timescales (Table~\ref{tab:100year-runtimes}). Efficiency gains compared to the other models are further emphasized in this long sequence setting, with SPF achieving much faster results than all probabilistic models on both timescales except for Multi-Yearly flow at the yearly timescale, for which it nearly matches in runtime. Importantly, SPF can generate 100-year probabilistic samples of monthly data in 1-2 minutes, which is much faster than previous weather-scale autoregressive models (which take nearly 3 hours) and massively faster than the physical simulations (which take weeks to months). 

\subsection{Other Results}
\begin{figure*}
    \centering
    \includegraphics[width=0.95\linewidth]{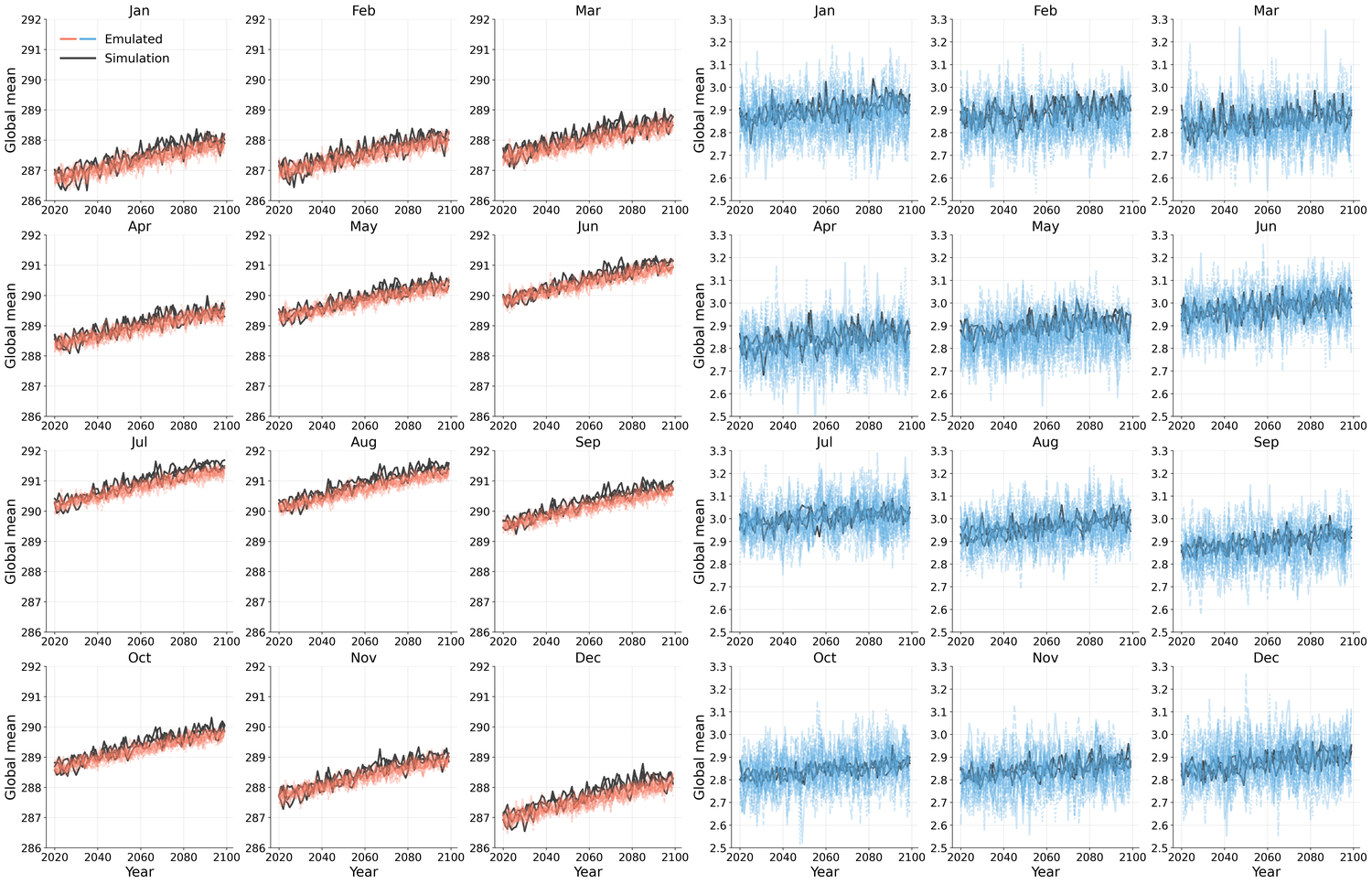}
    \caption{\textbf{Monthly latitude-weighted global means of the 200M SPF on ClimateBench SSP2-4.5.} Temperature shown in the left three columns (red) and precipitation in the right three columns (blue).}
    \label{fig:monthly_global_mean}
\end{figure*}

\begin{figure*}
    \centering
    \includegraphics[width=0.95\linewidth]{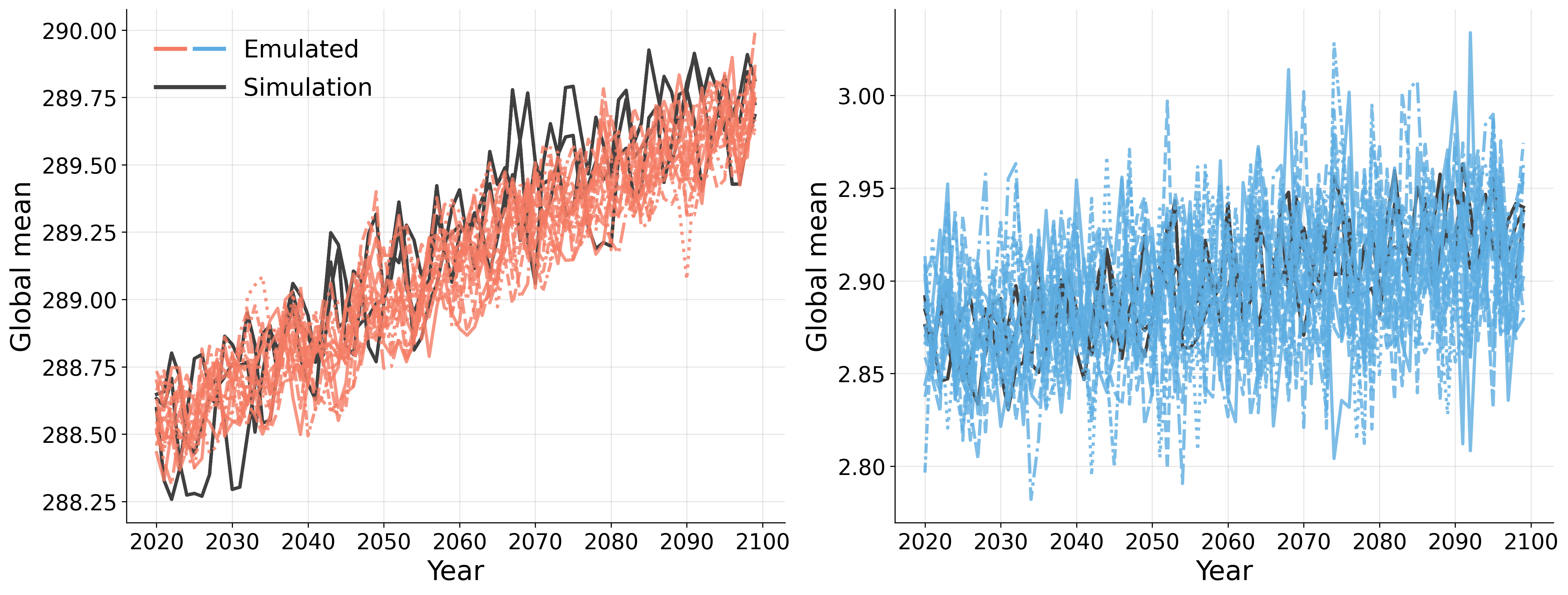}
    \caption{\textbf{Yearly latitude-weighted global means of the 200M SPF on ClimateBench SSP2-4.5.} Temperature shown in the left column (red) and precipitation in the right column (blue).}
    \label{fig:yearly_global_mean}
\end{figure*}

\begin{figure*}
    \centering
    \includegraphics[width=0.95\linewidth]{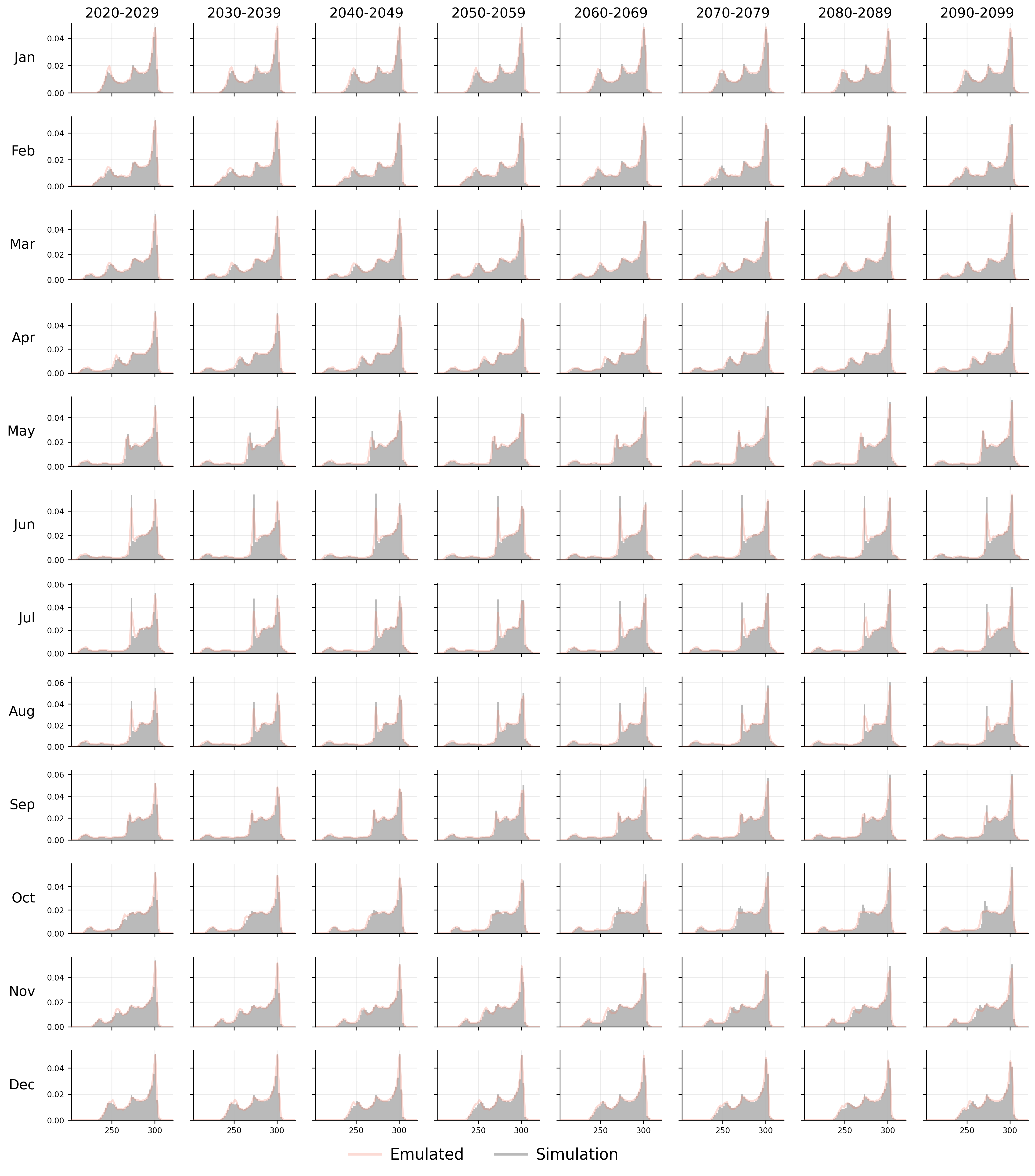}
    \caption{\textbf{Monthly temperature histograms of the 200M SPF model on ClimateBench SSP2-4.5, broken down by month and decade.}}
    \label{fig:hist_temp}
\end{figure*}

\begin{figure*}
    \centering
    \includegraphics[width=0.95\linewidth]{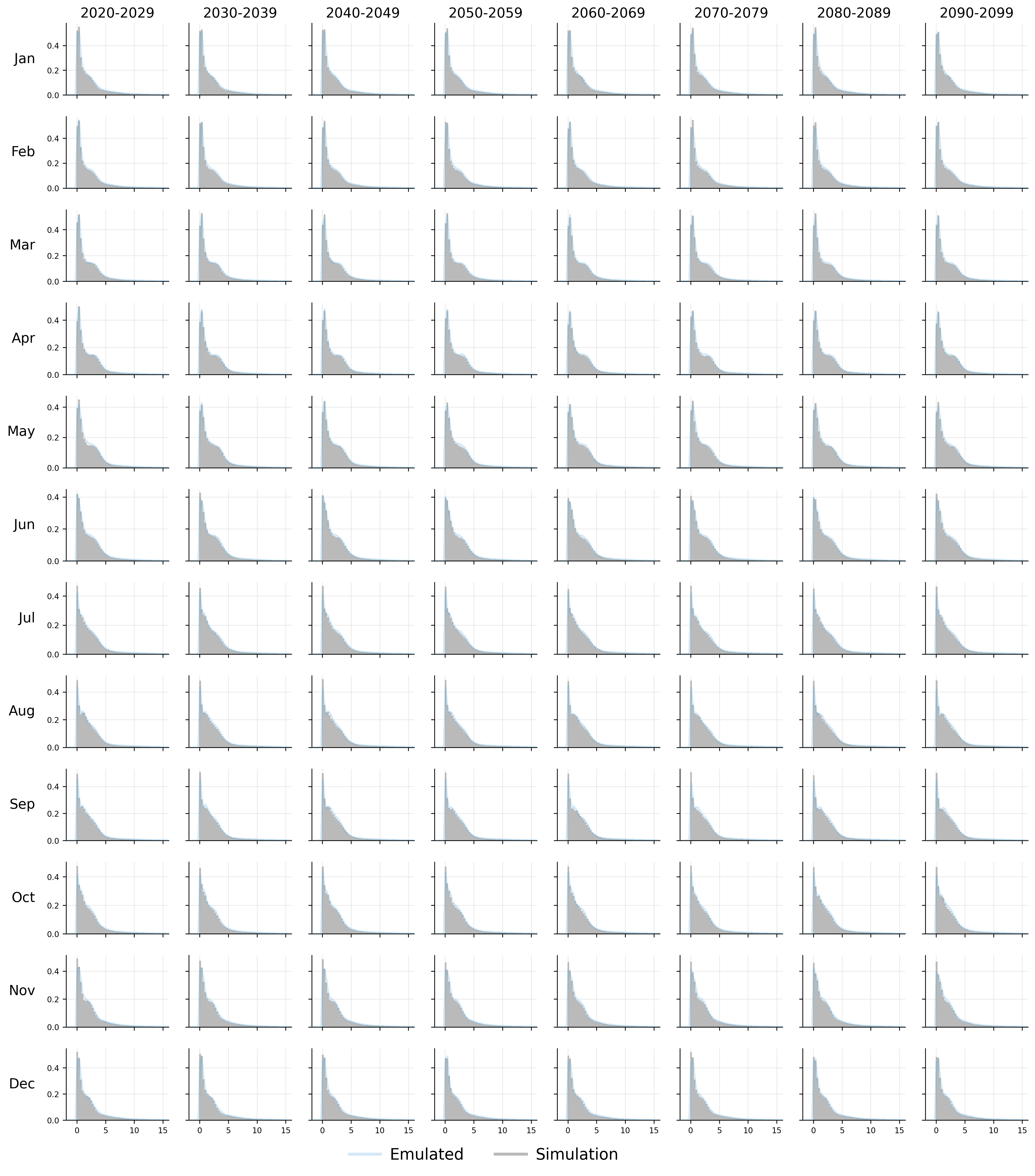}
    \caption{\textbf{Monthly precipitation histograms of the 200M SPF model on ClimateBench SSP2-4.5, broken down by month and decade.}}
    \label{fig:hist_precip}
\end{figure*}

\begin{figure*}
    \centering
    \includegraphics[width=0.95\linewidth]{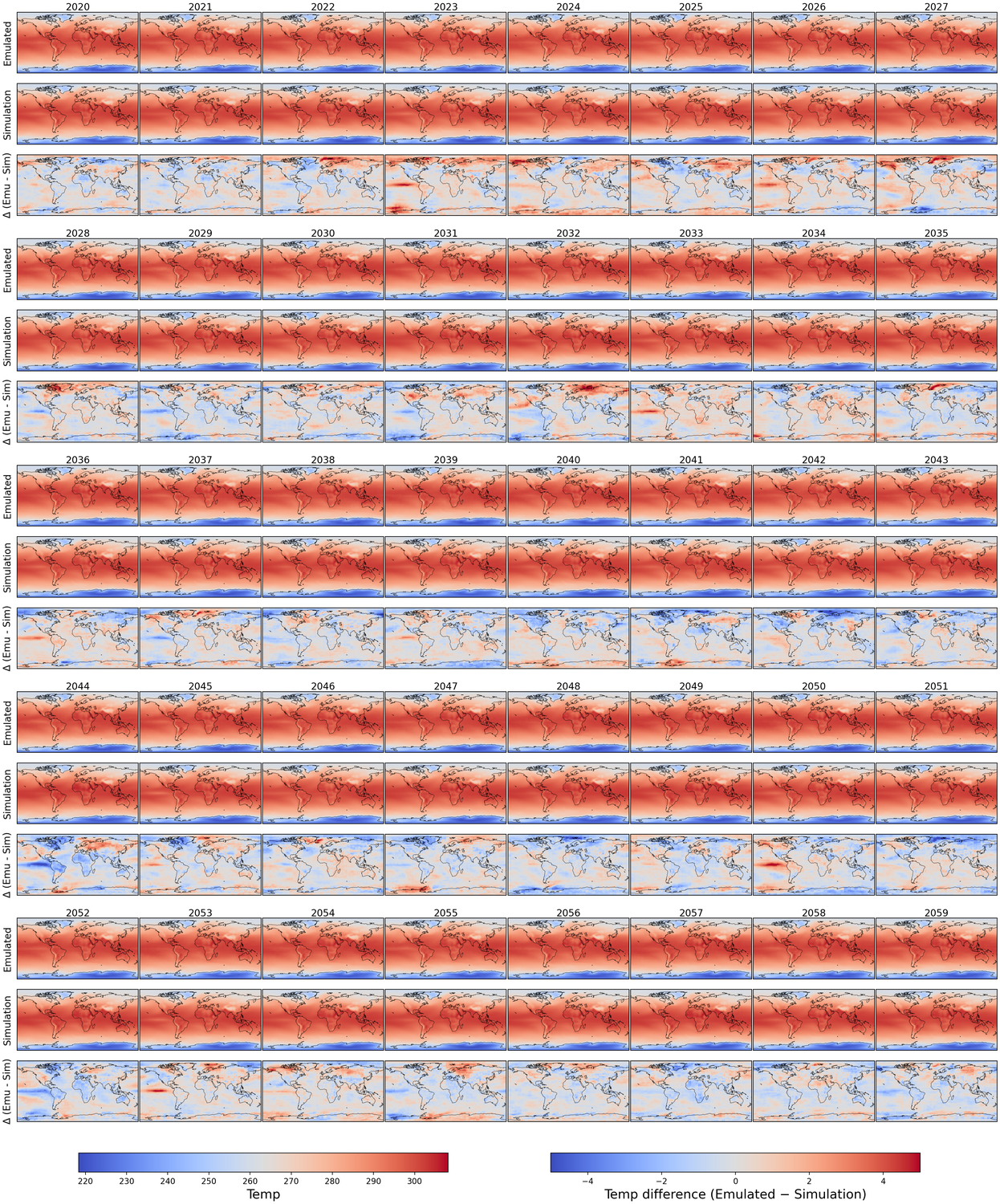}
    \caption{\textbf{ClimateBench samples of yearly emulated, simulated, and difference temperature maps (2020–2059, SSP2-4.5).}}
    \label{fig:yearly_samples_2020_tas}
\end{figure*}

\begin{figure*}
    \centering
    \includegraphics[width=0.95\linewidth]{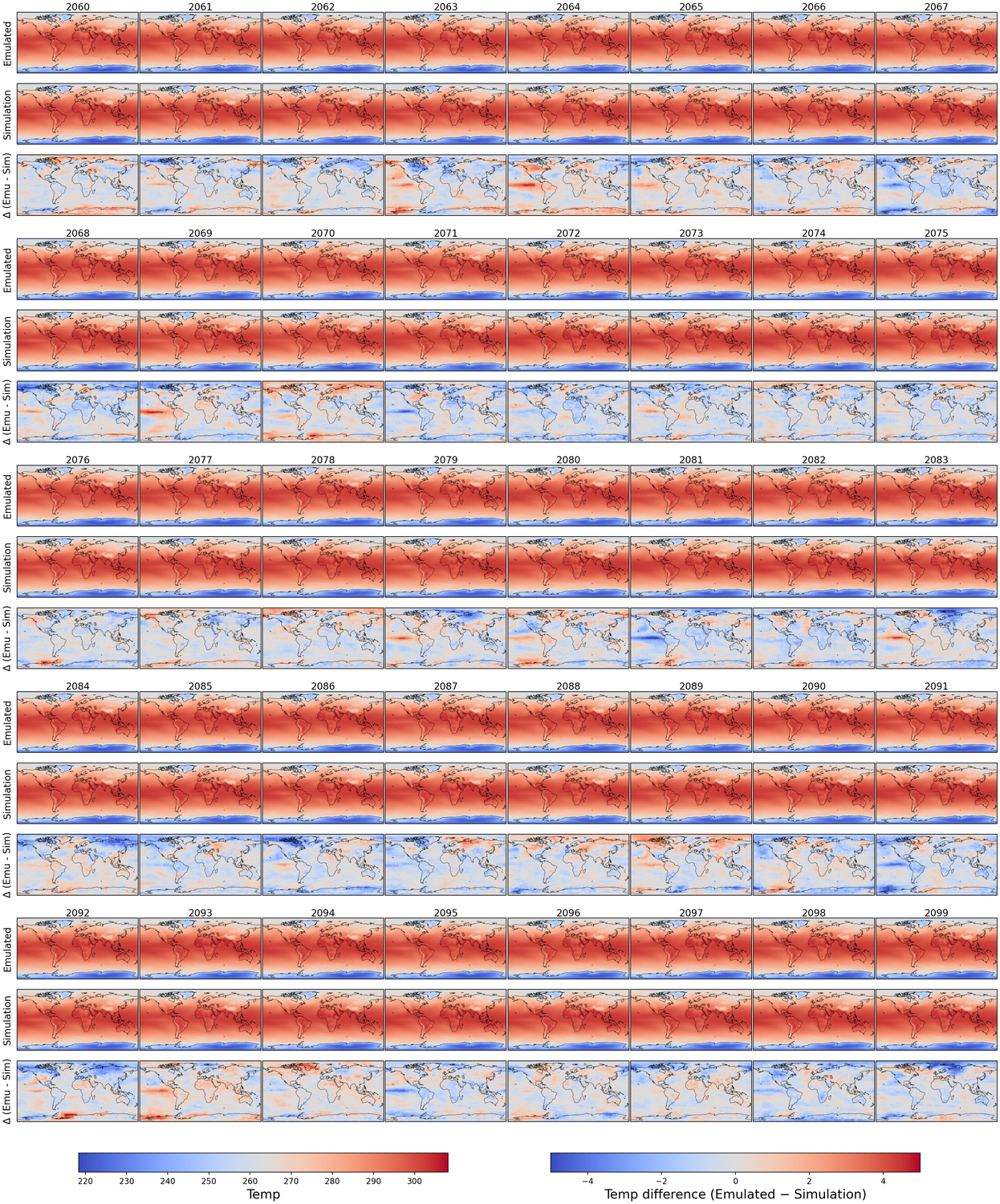}
    \caption{\textbf{ClimateBench samples of yearly emulated, simulated, and difference temperature maps (2060–2099, SSP2-4.5).}}
    \label{fig:yearly_samples_2060_tas}
\end{figure*}

\begin{figure*}
    \centering
    \includegraphics[width=0.95\linewidth]{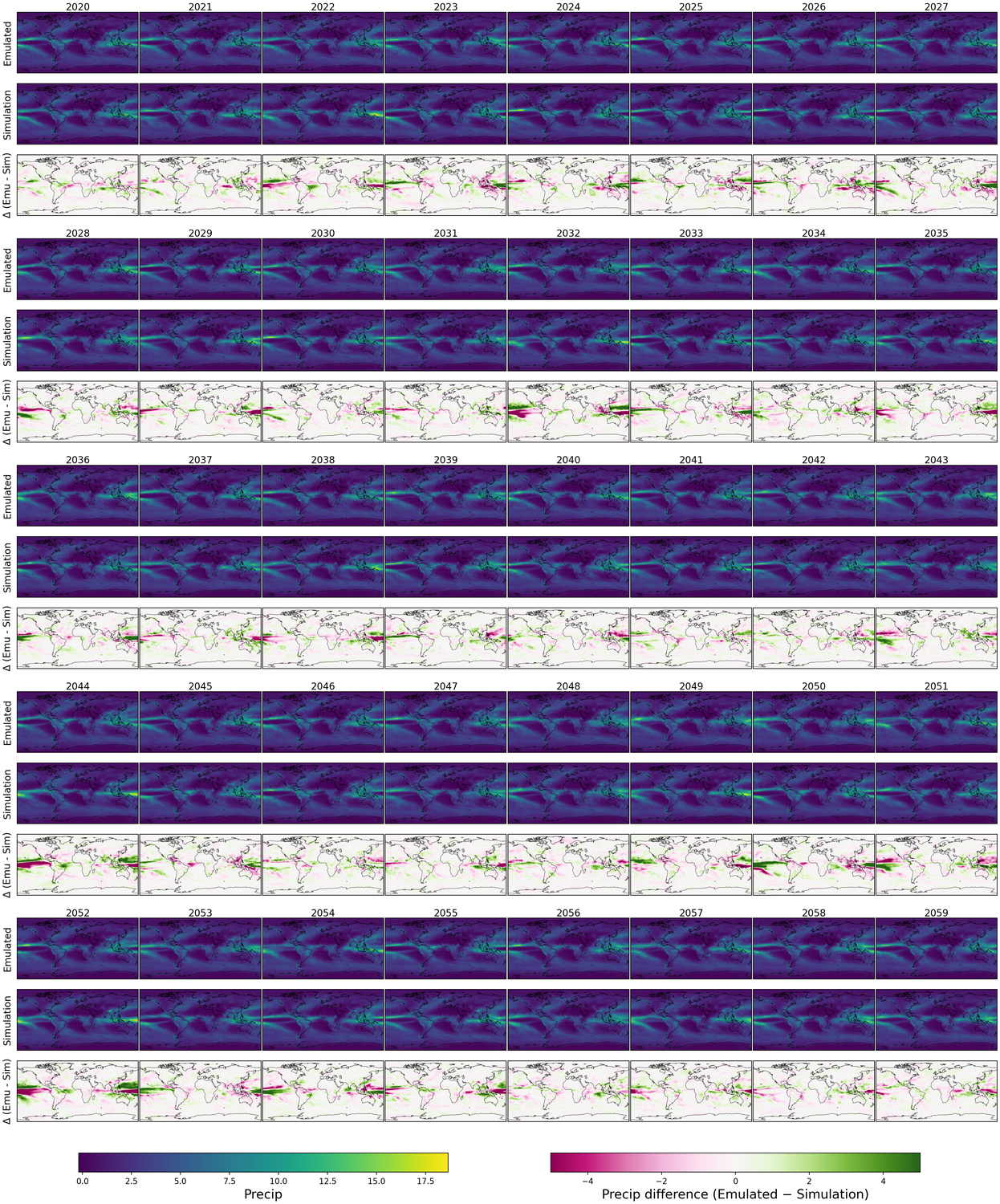}
    \caption{\textbf{ClimateBench samples of yearly emulated, simulated, and difference precipitation maps (2020–2059, SSP2-4.5).}}
    \label{fig:yearly_samples_2020_pr}
\end{figure*}

\begin{figure*}
    \centering
    \includegraphics[width=0.95\linewidth]{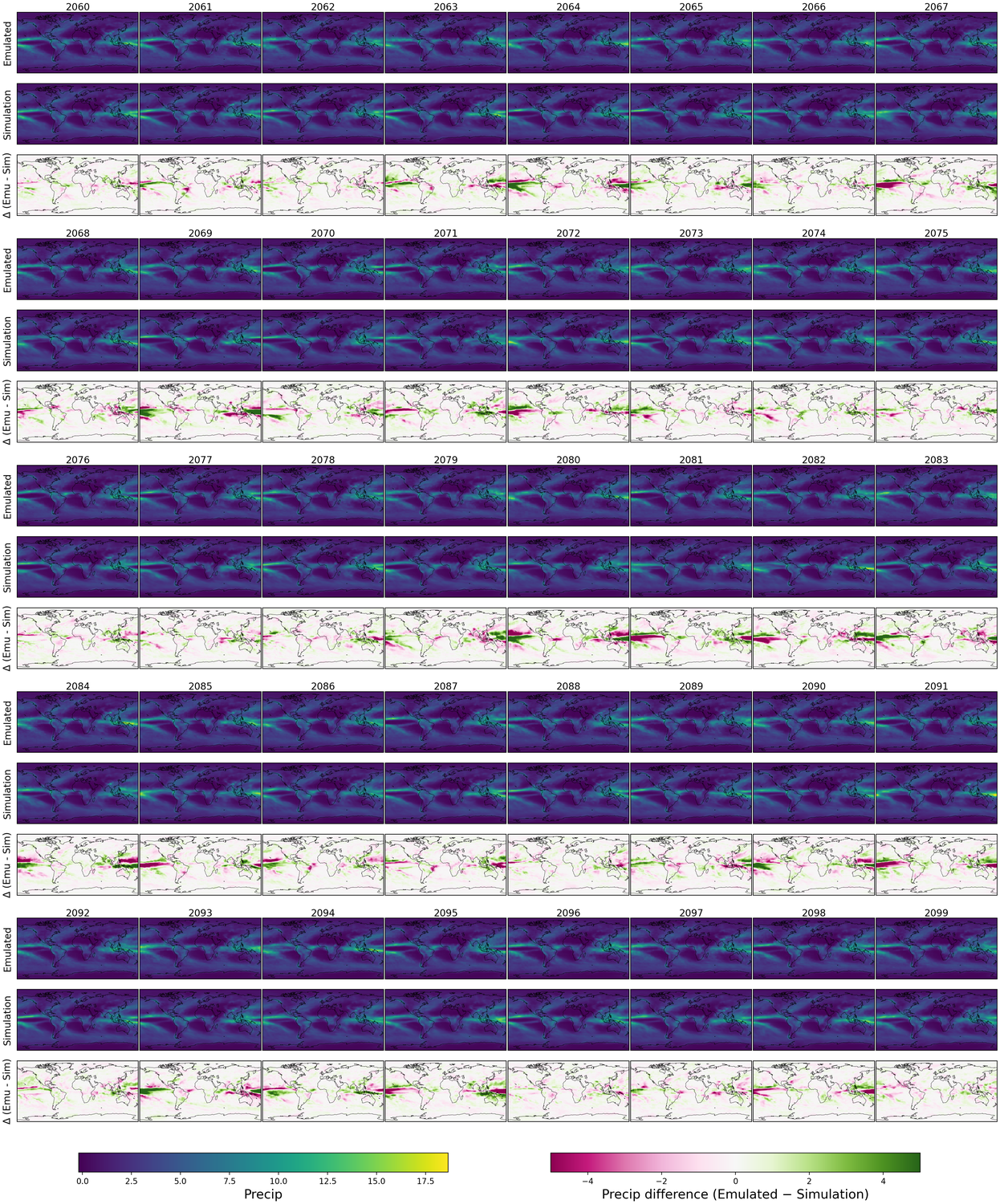}
    \caption{\textbf{ClimateBench samples of yearly emulated, simulated, and difference precipitation maps (2060–2099, SSP2-4.5).}}
    \label{fig:yearly_samples_2060_pr}
\end{figure*}

\begin{figure*}
    \centering
    \includegraphics[width=0.98\linewidth]{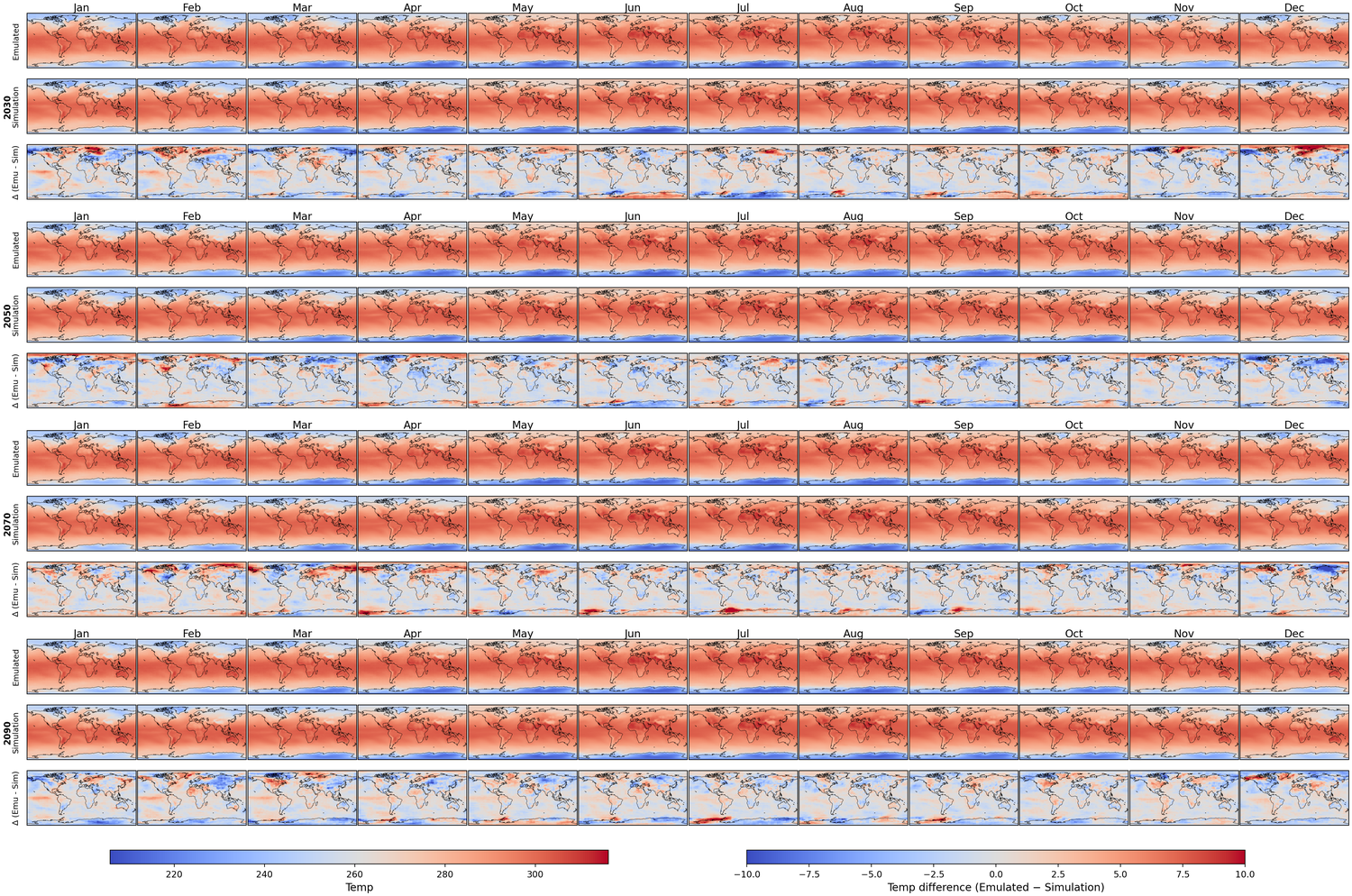}
    \caption{\textbf{ClimateBench samples of monthly emulated, simulated, and difference temperature maps (SSP2-4.5).}}
    \label{fig:monthly_samples_tas}
\end{figure*}

\begin{figure*}
    \centering
    \includegraphics[width=0.98\linewidth]{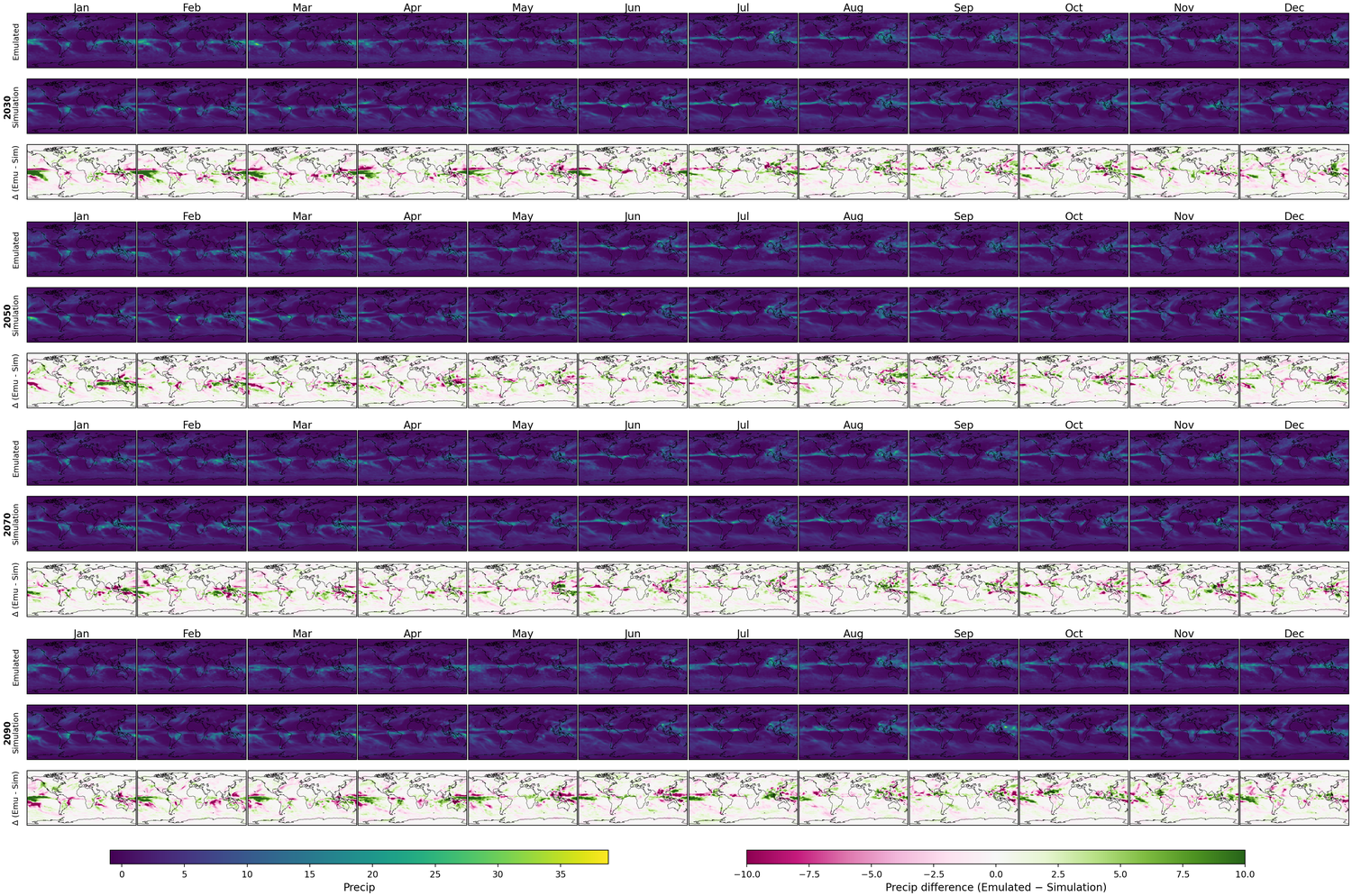}
    \caption{\textbf{ClimateBench samples of monthly emulated, simulated, and difference precipitation maps (SSP2-4.5).}}
    \label{fig:monthly_samples_pr}
\end{figure*}

\begin{figure*}
    \centering
    \includegraphics[width=0.5\linewidth]{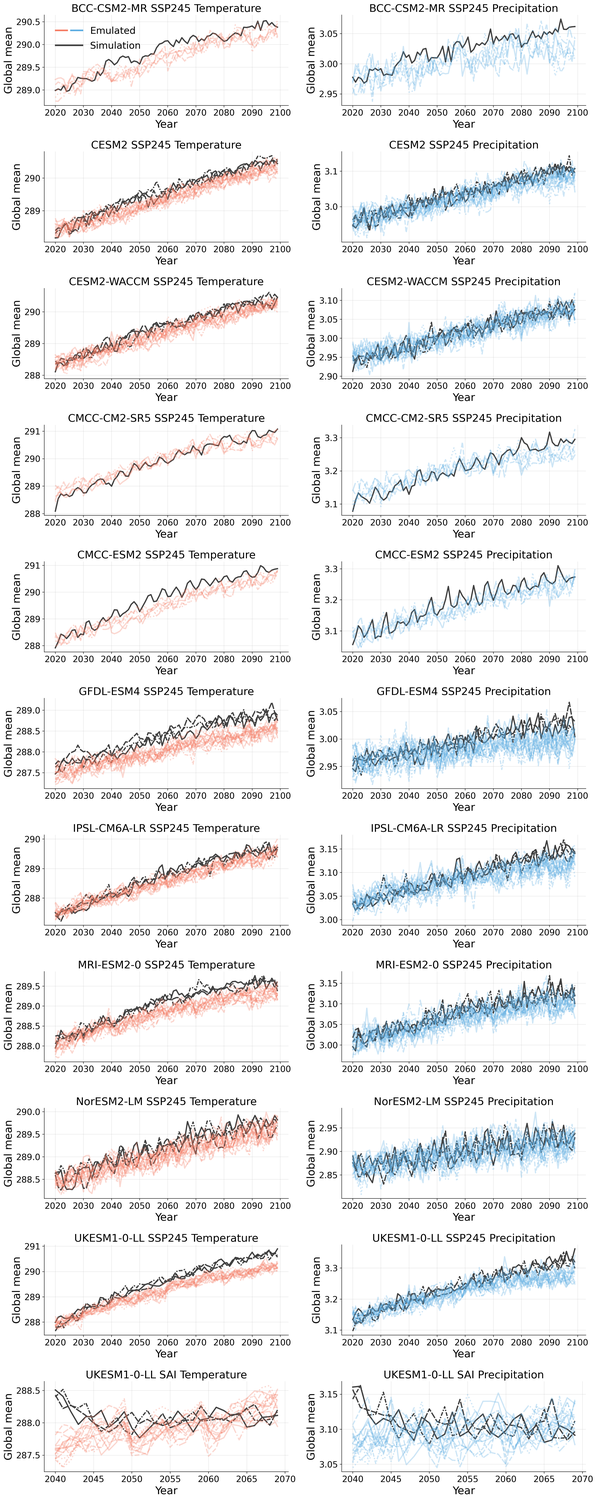}
    \caption{\textbf{Yearly latitude-weighted global means of the 600M SPF model on SSP2-4.5 across the 10 climate models and the SAI experiment on UKESM1-0-LL in ClimateSuite.}22}
    \label{fig:multi_yearly_global_mean}
\end{figure*}

\begin{figure*}
    \centering
    \includegraphics[width=0.95\linewidth]{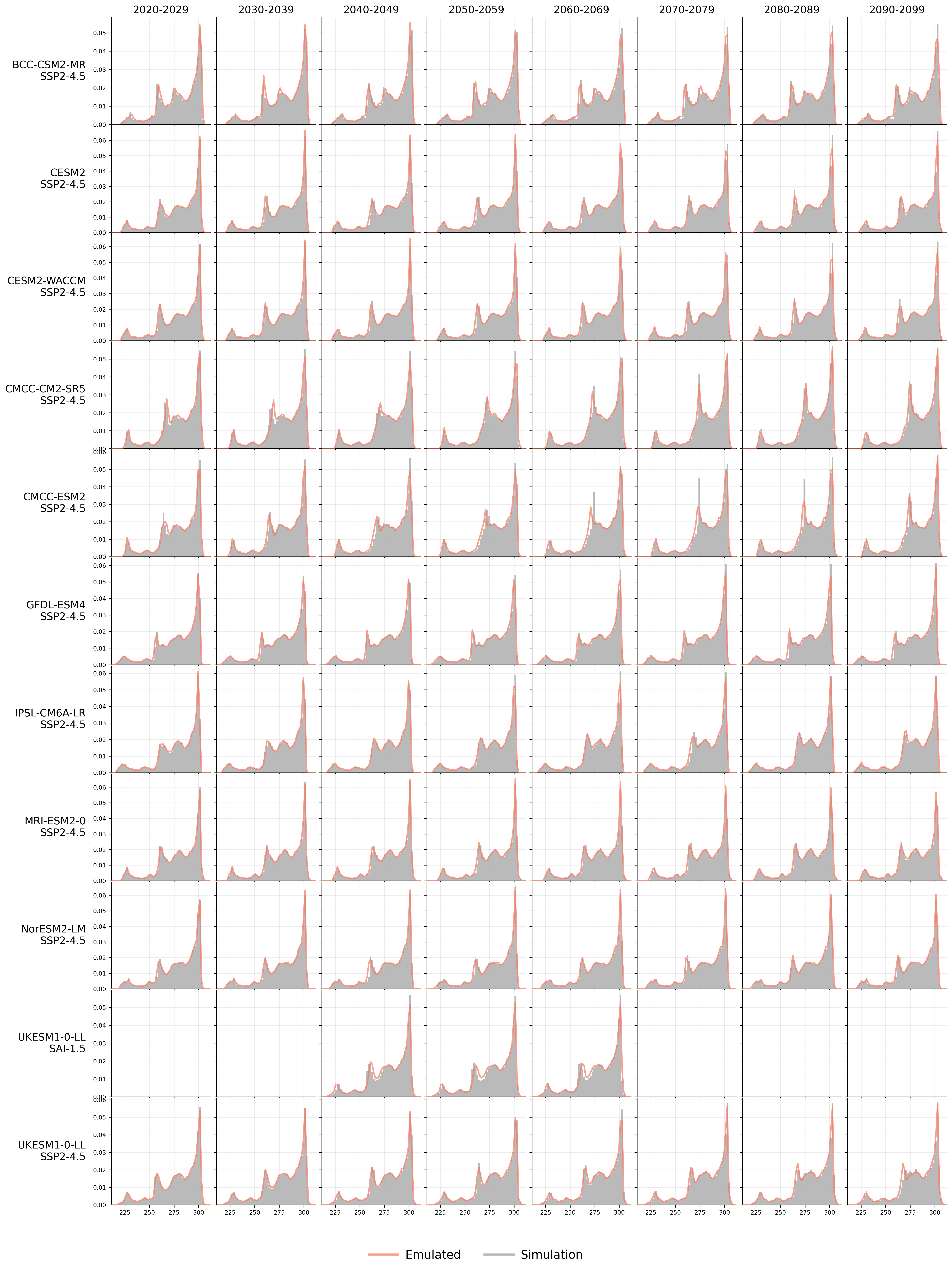}
    \caption{\textbf{Yearly temperature histograms of the 600M SPF model on SSP2-4.5 across the 10 climate models and the SAI experiment on UKESM1-0-LL, broken down by decade.}}
    \label{fig:multi_hist_temp}
\end{figure*}

\begin{figure*}
    \centering
    \includegraphics[width=0.95\linewidth]{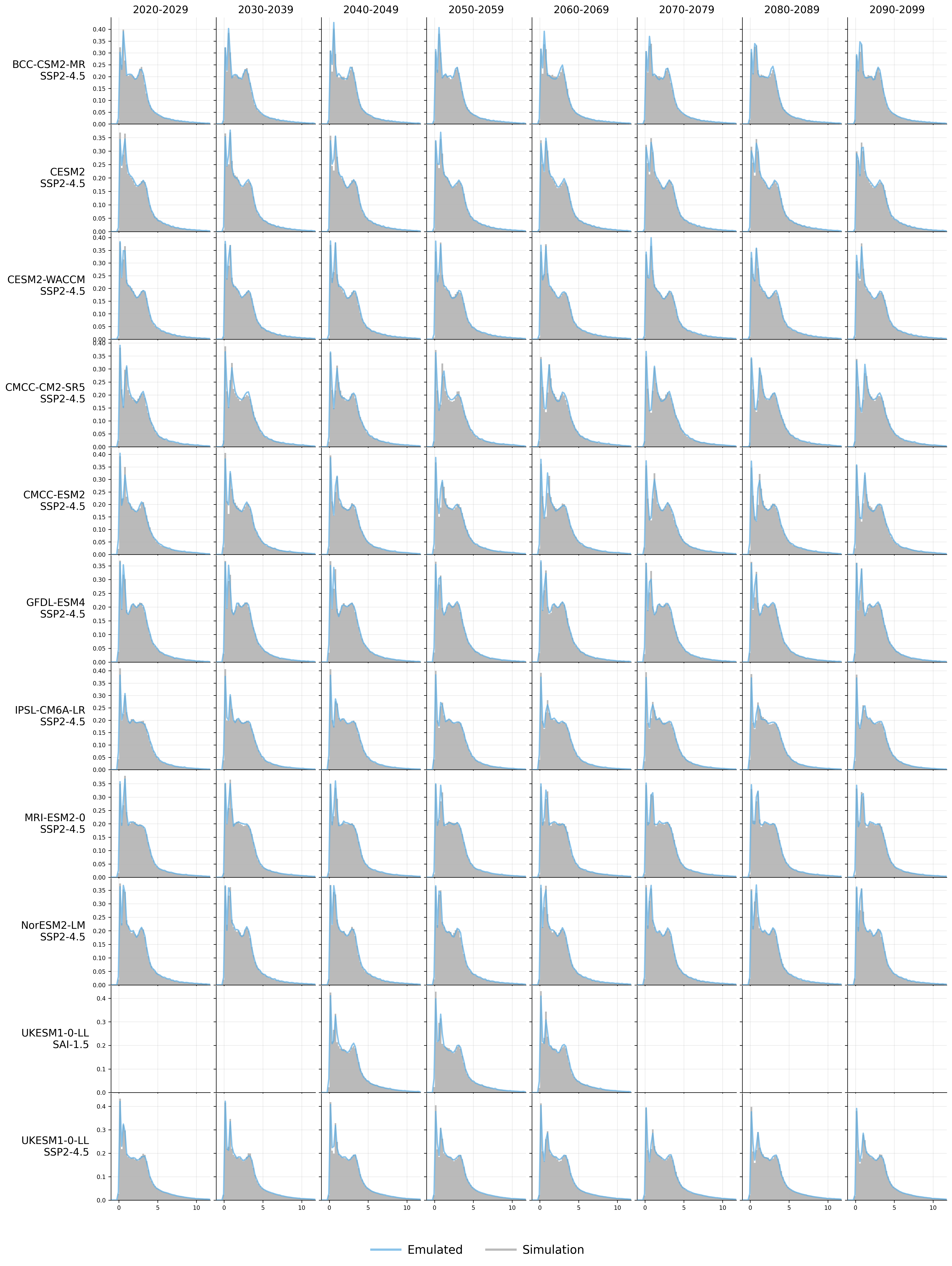}
    \caption{\textbf{Yearly precipitation histograms of the 600M SPF model on SSP2-4.5 across the 10 climate models and the SAI experiment on UKESM1-0-LL, broken down by decade.}}
    \label{fig:multi_hist_precip}
\end{figure*}

\begin{figure*}
    \centering
    \includegraphics[width=0.95\linewidth]{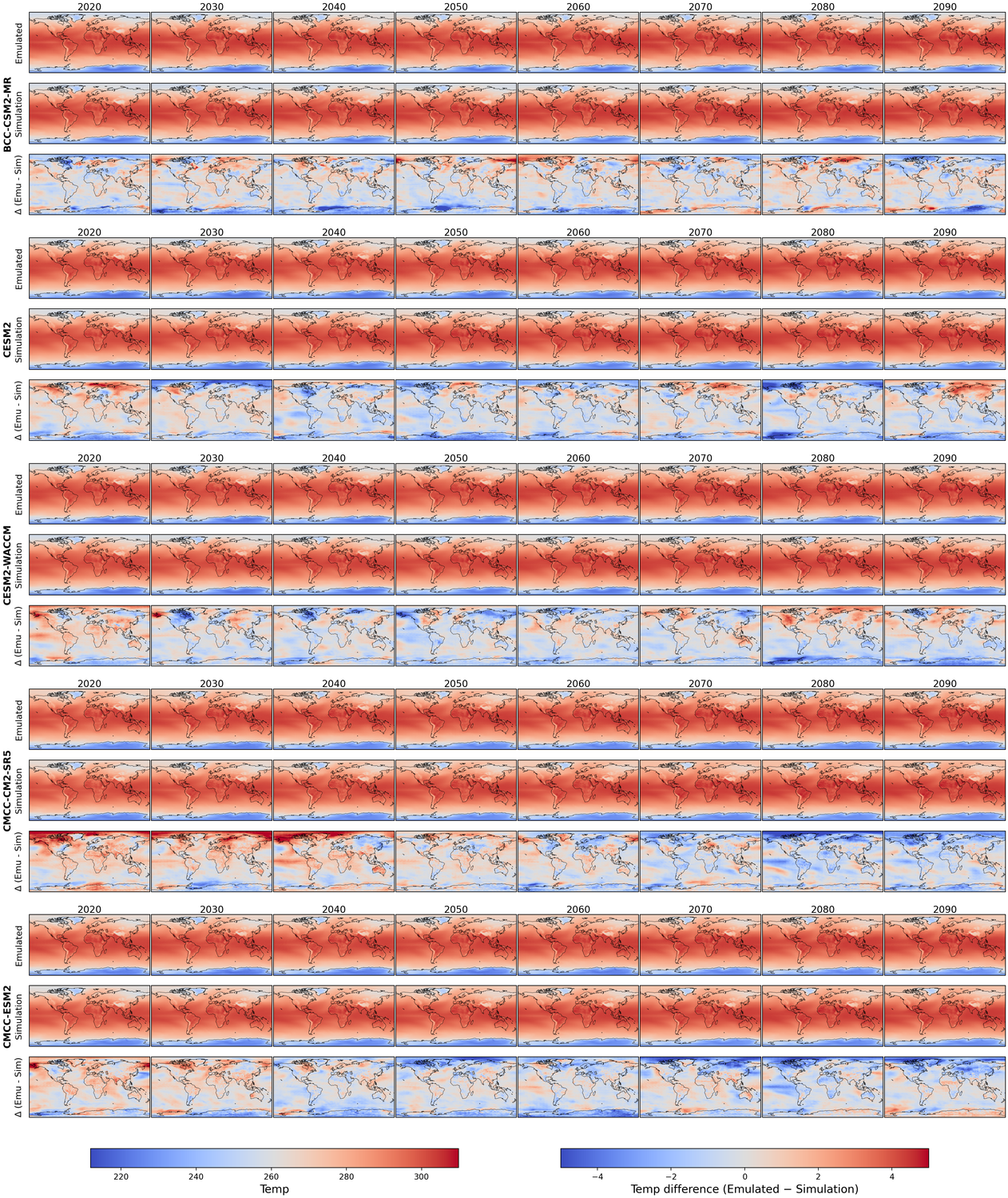}
    \caption{\textbf{Per-model ClimateSuite samples of yearly emulated, simulated, and difference temperature maps (SSP2-4.5).}}
    \label{fig:multi_yearly_samples_temp1}
\end{figure*}

\begin{figure*}
    \centering
    \includegraphics[width=0.95\linewidth]{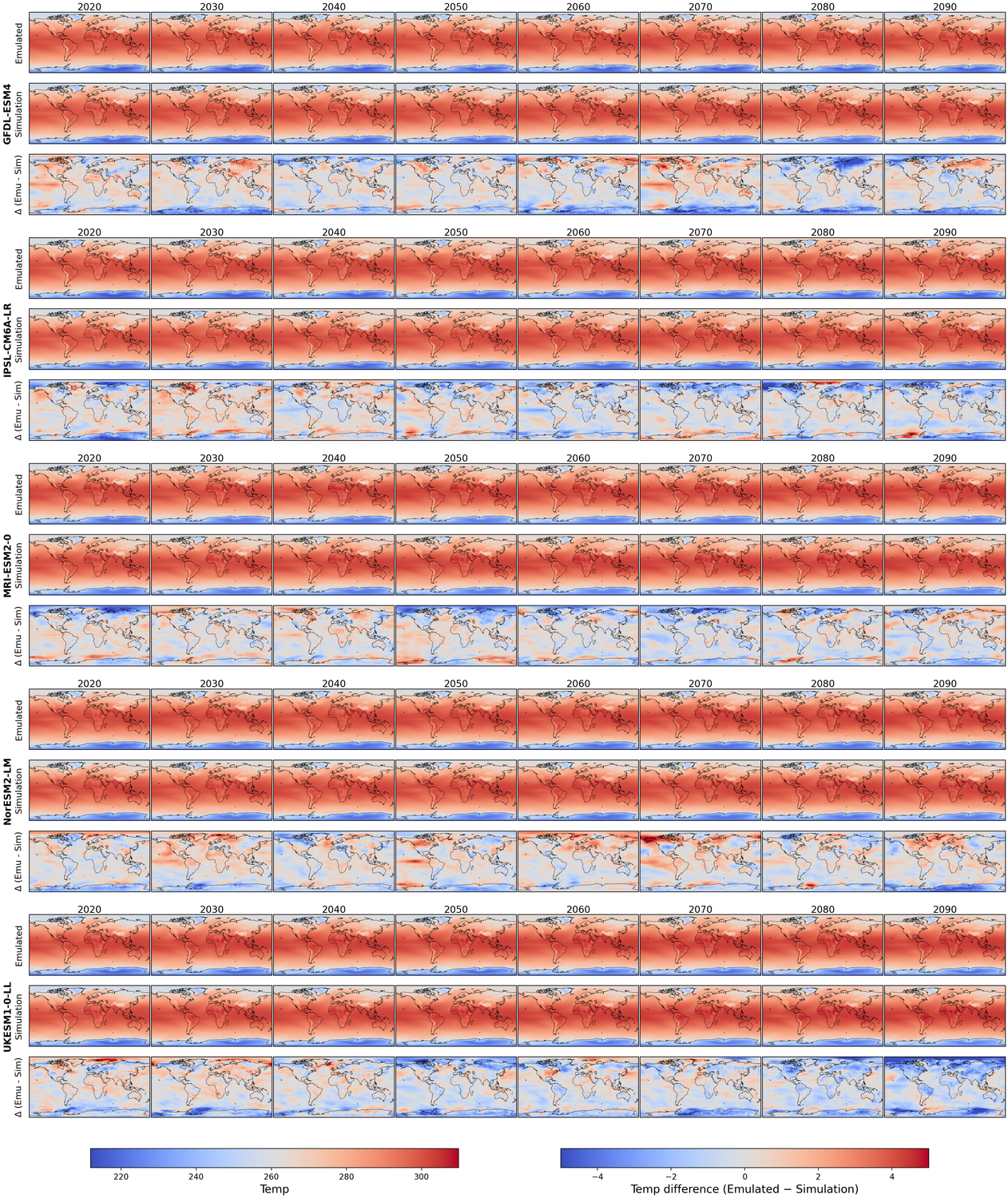}
    \caption{\textbf{Per-model ClimateSuite samples of yearly emulated, simulated, and difference temperature maps (SSP2-4.5).}}
    \label{fig:multi_yearly_samples_temp2}
\end{figure*}

\begin{figure*}
    \centering
    \includegraphics[width=0.95\linewidth]{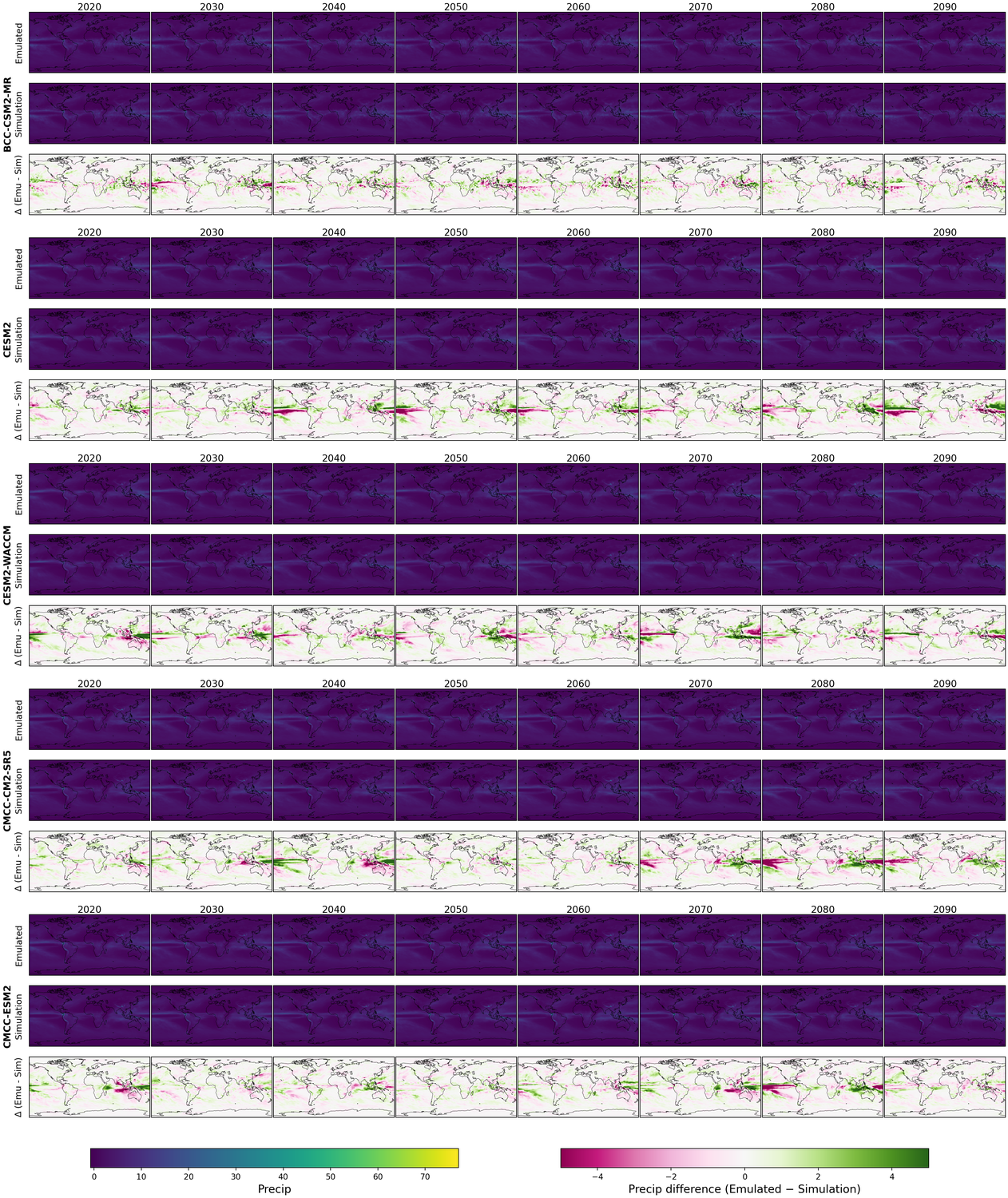}
    \caption{\textbf{Per-model ClimateSuite samples of yearly emulated, simulated, and difference precipitation maps (SSP2-4.5).}}
    \label{fig:multi_yearly_samples_precip1}
\end{figure*}

\begin{figure*}
    \centering
    \includegraphics[width=0.95\linewidth]{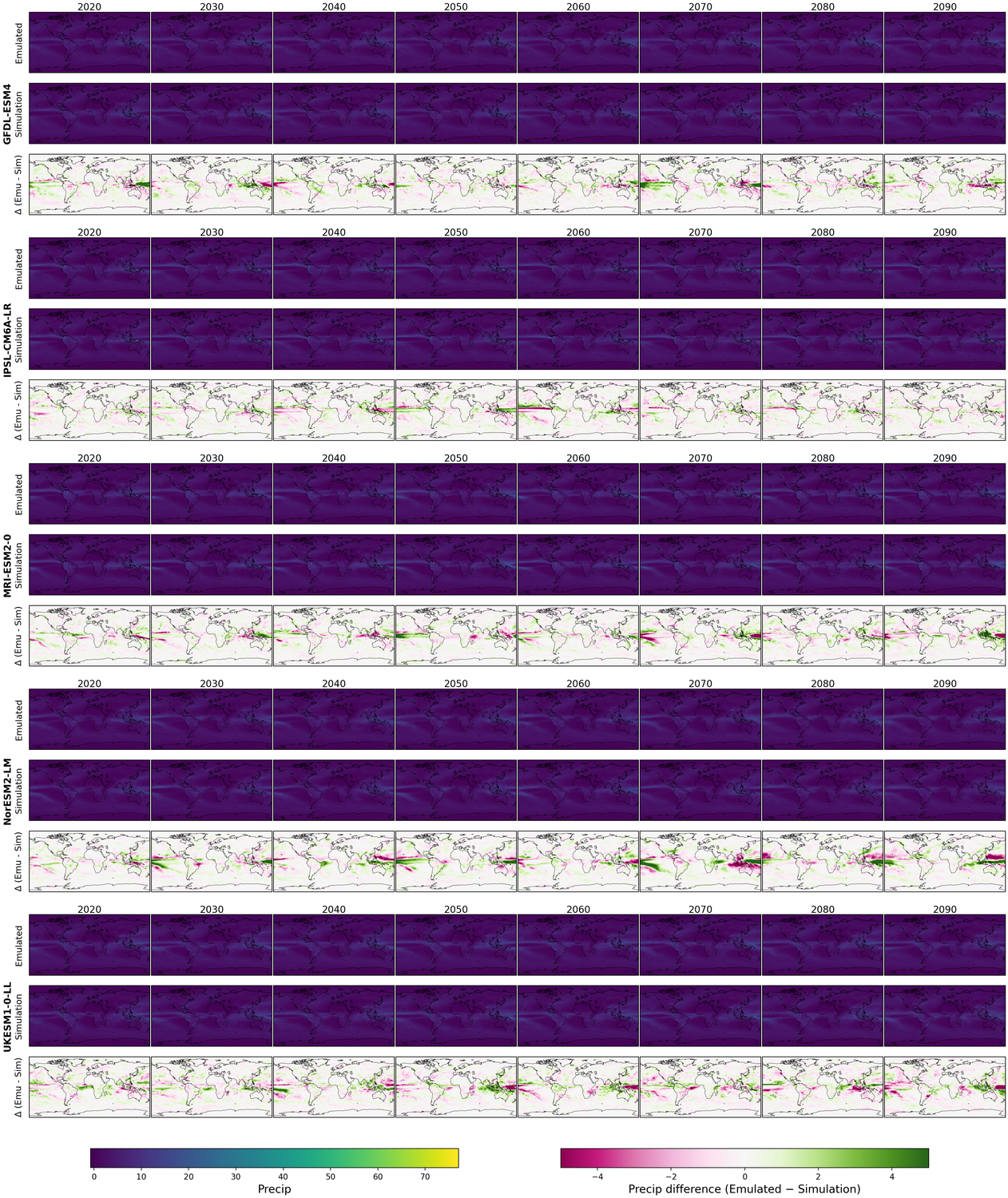}
    \caption{\textbf{Per-model ClimateSuite samples of yearly emulated, simulated, and difference precipitation maps (SSP2-4.5).}}
    \label{fig:multi_yearly_samples_precip2}
\end{figure*}

\begin{figure*}
    \centering
    \includegraphics[width=0.9\linewidth]{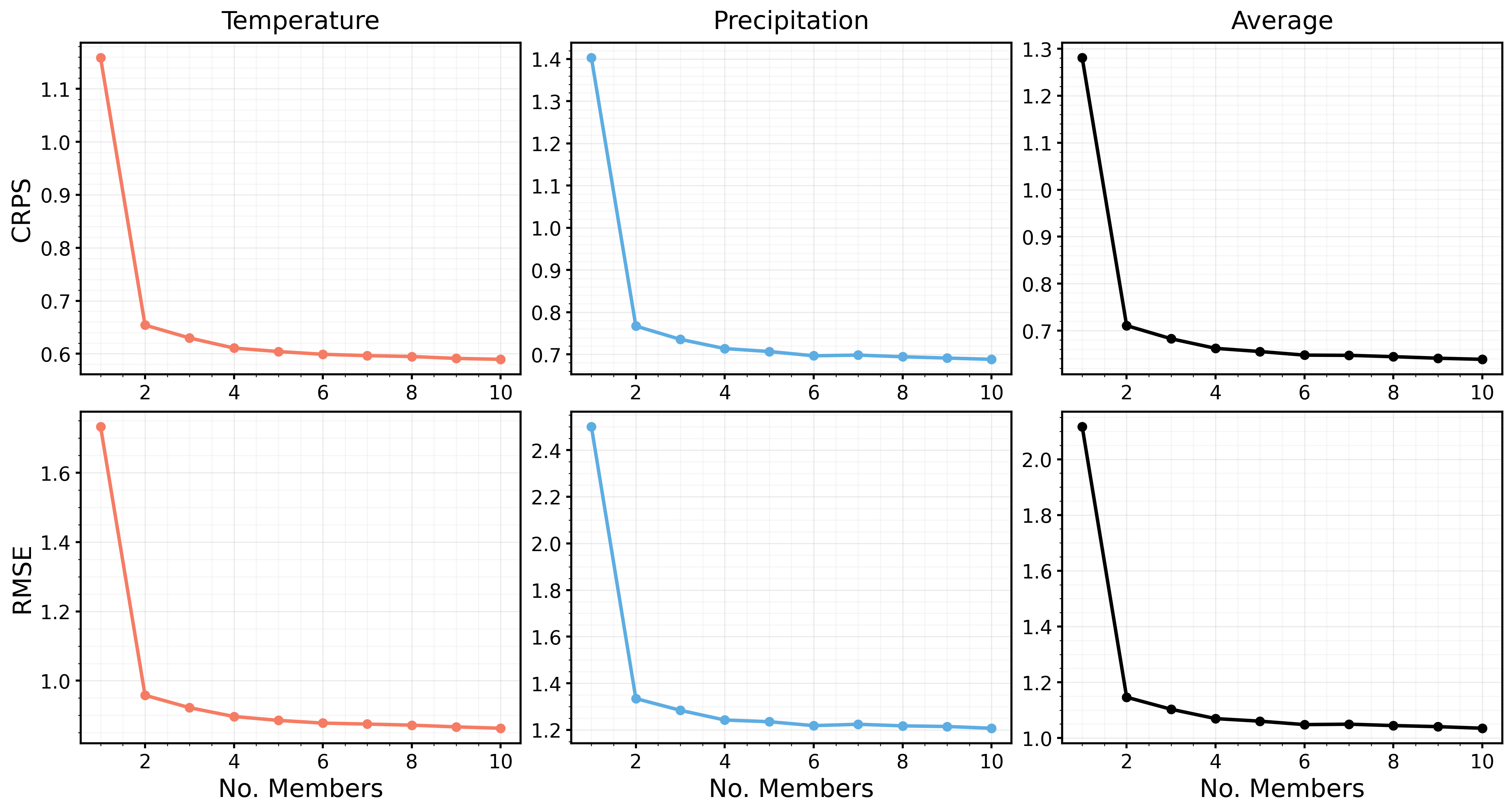}
    \caption{\textbf{Effect of number of members on SPF performance on ClimateBench SSP2-4.5.}}
    \label{fig:member_ablation}
\end{figure*}

We report a variety of additional results, including tuning ClimaX settings (Table~\ref{tab:climax}), global means of SPF compared to simulations in ClimateBench (Figures~\ref{fig:monthly_global_mean}--\ref{fig:yearly_global_mean}), SPF histograms on ClimateBench (Figures~\ref{fig:hist_temp}--\ref{fig:hist_precip}), SPF samples on ClimateBench (Figures~\ref{fig:yearly_samples_2020_tas}--\ref{fig:monthly_samples_pr}), multi-model global means of SPF compared to simulations in ClimateSuite (Figures~\ref{fig:multi_yearly_global_mean}), SPF histograms on ClimateSuite (Figures~\ref{fig:multi_hist_temp}--\ref{fig:multi_hist_precip}), SPF samples of multiple models in ClimateSuite (Figures~\ref{fig:multi_yearly_samples_temp1}--\ref{fig:multi_yearly_samples_precip2}), and an ablation measuring the impact of using variable number of ensemble members on SPF performance on ClimateBench (Figure~\ref{fig:member_ablation}).

\begin{table}[!]
\centering
\begin{tabular}{lccc}
\toprule
Procedure & Resolution & Patch  Size & RMSE\\
\midrule
Fine-Tuned & High & 16 & 0.546\\
Fine-Tuned & High & 4 & 0.556\\
Fine-Tuned & Low & 2 & 0.577\\
Fine-Tuned & Low & 16 & 0.594\\
Frozen & High & 16 & 0.619\\
Frozen & Low & 16 & 0.721\\
Frozen & High & 4 & 0.721\\
Frozen & Low & 2 & 0.811\\
\bottomrule
\end{tabular}
\caption{\textbf{ClimateBench yearly temporal resolution results under different ClimaX variants.}}
\label{tab:climax}
\end{table}

%% file: main.bib
@String(AAAI = {AAAI})

@article{jin2024pyramidal,
  title={Pyramidal flow matching for efficient video generative modeling},
  author={Jin, Yang and Sun, Zhicheng and Li, Ningyuan and Xu, Kun and Jiang, Hao and Zhuang, Nan and Huang, Quzhe and Song, Yang and Mu, Yadong and Lin, Zhouchen},
  journal={arXiv preprint arXiv:2410.05954},
  year={2024}
}

@article{chen2025pixelflow,
  title={PixelFlow: Pixel-Space Generative Models with Flow},
  author={Chen, Shoufa and Ge, Chongjian and Zhang, Shilong and Sun, Peize and Luo, Ping},
  journal={arXiv preprint arXiv:2504.07963},
  year={2025}
}

@article{campbell2023trans,
  title={Trans-dimensional generative modeling via jump diffusion models},
  author={Campbell, Andrew and Harvey, William and Weilbach, Christian and De Bortoli, Valentin and Rainforth, Thomas and Doucet, Arnaud},
  journal={Advances in Neural Information Processing Systems},
  volume={36},
  pages={42217--42257},
  year={2023}
}

@article{yan2024perflow,
  title={Perflow: Piecewise rectified flow as universal plug-and-play accelerator},
  author={Yan, Hanshu and Liu, Xingchao and Pan, Jiachun and Liew, Jun Hao and Liu, Qiang and Feng, Jiashi},
  journal={arXiv preprint arXiv:2405.07510},
  year={2024}
}

@article{elfwing2018sigmoid,
  title={Sigmoid-weighted linear units for neural network function approximation in reinforcement learning},
  author={Elfwing, Stefan and Uchibe, Eiji and Doya, Kenji},
  journal={Neural networks},
  volume={107},
  pages={3--11},
  year={2018},
  publisher={Elsevier}
}

@inproceedings{sohl2015deep,
  title={Deep unsupervised learning using nonequilibrium thermodynamics},
  author={Sohl-Dickstein, Jascha and Weiss, Eric and Maheswaranathan, Niru and Ganguli, Surya},
  booktitle={International conference on machine learning},
  pages={2256--2265},
  year={2015},
  organization={pmlr}
}

@article{song2019generative,
  title={Generative modeling by estimating gradients of the data distribution},
  author={Song, Yang and Ermon, Stefano},
  journal={Advances in neural information processing systems},
  volume={32},
  year={2019}
}

@article{ho2020denoising,
  title={Denoising diffusion probabilistic models},
  author={Ho, Jonathan and Jain, Ajay and Abbeel, Pieter},
  journal={Advances in neural information processing systems},
  volume={33},
  pages={6840--6851},
  year={2020}
}

@article{lipman2022flow,
  title={Flow matching for generative modeling},
  author={Lipman, Yaron and Chen, Ricky TQ and Ben-Hamu, Heli and Nickel, Maximilian and Le, Matt},
  journal={arXiv preprint arXiv:2210.02747},
  year={2022}
}

@inproceedings{onken2021ot,
  title={Ot-flow: Fast and accurate continuous normalizing flows via optimal transport},
  author={Onken, Derek and Fung, Samy Wu and Li, Xingjian and Ruthotto, Lars},
  booktitle={Proceedings of the AAAI Conference on Artificial Intelligence},
  volume={35},
  number={10},
  pages={9223--9232},
  year={2021}
}

@article{papamakarios2021normalizing,
  title={Normalizing flows for probabilistic modeling and inference},
  author={Papamakarios, George and Nalisnick, Eric and Rezende, Danilo Jimenez and Mohamed, Shakir and Lakshminarayanan, Balaji},
  journal={Journal of Machine Learning Research},
  volume={22},
  number={57},
  pages={1--64},
  year={2021}
}

@article{liu2022flow,
  title={Flow straight and fast: Learning to generate and transfer data with rectified flow},
  author={Liu, Xingchao and Gong, Chengyue and Liu, Qiang},
  journal={arXiv preprint arXiv:2209.03003},
  year={2022}
}

@article{ho2022cascaded,
  title={Cascaded diffusion models for high fidelity image generation},
  author={Ho, Jonathan and Saharia, Chitwan and Chan, William and Fleet, David J and Norouzi, Mohammad and Salimans, Tim},
  journal={Journal of Machine Learning Research},
  volume={23},
  number={47},
  pages={1--33},
  year={2022}
}

@article{wang2025lavie,
  title={Lavie: High-quality video generation with cascaded latent diffusion models},
  author={Wang, Yaohui and Chen, Xinyuan and Ma, Xin and Zhou, Shangchen and Huang, Ziqi and Wang, Yi and Yang, Ceyuan and He, Yinan and Yu, Jiashuo and Yang, Peiqing and others},
  journal={International Journal of Computer Vision},
  volume={133},
  number={5},
  pages={3059--3078},
  year={2025},
  publisher={Springer}
}

@misc{ho2022imagen,
  title        = {Imagen Video: High Definition Video Generation with Diffusion Models},
  author       = {Ho, Jonathan and Chan, William and Saharia, Chitwan and Whang, Jay and Gao, Ruiqi and Gritsenko, Alexey and Kingma, Diederik P. and Poole, Ben and Norouzi, Mohammad and Fleet, David J. and Salimans, Tim},
  year         = {2022},
  eprint       = {2210.02303},
  archivePrefix= {arXiv},
  primaryClass = {cs.CV},
  note         = {arXiv preprint arXiv:2210.02303},
  url          = {https://arxiv.org/abs/2210.02303},
}

@inproceedings{cachay2024sphericaldyffusion,
  author={Salva R{\"u}hling Cachay and Brian Henn and Oliver Watt-Meyer and Christopher S. Bretherton and Rose Yu},
  title={Probabilistic Emulation of a Global Climate Model with Spherical {DYffusion}},
  booktitle={Advances in Neural Information Processing Systems (NeurIPS)},
  year={2024}
}

@article{wattmeyer2024ace,
  author={Oliver Watt-Meyer and Gideon Dresdner and Jeremy McGibbon and Spencer K. Clark and Brian Henn and James Duncan and Noah D. Brenowitz and Karthik Kashinath and Michael S. Pritchard and Boris Bonev and Matthew E. Peters and Christopher S. Bretherton},
  title={{ACE}: A fast, skillful learned global atmospheric model for climate prediction},
  journal={arXiv preprint arXiv:2310.02074},
  year={2024}
}

@article{watsonparris2022climatebench,
  author={Duncan Watson-Parris and Y. Rao and Delphine Olivi{\'e} and {\O}yvind Seland and Peer Nowack and Gustau Camps-Valls and Philip Stier and et al.},
  title={{ClimateBench} v1.0: a benchmark for data-driven climate projections},
  journal={Journal of Advances in Modeling Earth Systems},
  volume={14},
  number={10},
  pages={e2021MS002954},
  year={2022},
  doi={10.1029/2021MS002954}
}

@article{o2016scenario,
  title={The scenario model intercomparison project (ScenarioMIP) for CMIP6},
  author={O'Neill, Brian C and Tebaldi, Claudia and Van Vuuren, Detlef P and Eyring, Veronika and Friedlingstein, Pierre and Hurtt, George and Knutti, Reto and Kriegler, Elmar and Lamarque, Jean-Francois and Lowe, Jason and others},
  journal={Geoscientific Model Development},
  volume={9},
  number={9},
  pages={3461--3482},
  year={2016},
  publisher={Copernicus GmbH}
}

@article{kravitz2015geoengineering,
  title={The geoengineering model intercomparison project phase 6 (GeoMIP6): Simulation design and preliminary results},
  author={Kravitz, Ben and Robock, Alan and Tilmes, Simone and Boucher, Olivier and English, Jason M and Irvine, Peter J and Jones, Andy and Lawrence, Mark G and MacCracken, Michael and Muri, Helene and others},
  journal={Geoscientific Model Development},
  volume={8},
  number={10},
  pages={3379--3392},
  year={2015},
  publisher={Copernicus GmbH}
}

@article{eyring2016overview,
  title={Overview of the Coupled Model Intercomparison Project Phase 6 (CMIP6) experimental design and organization},
  author={Eyring, Veronika and Bony, Sandrine and Meehl, Gerald A and Senior, Catherine A and Stevens, Bjorn and Stouffer, Ronald J and Taylor, Karl E},
  journal={Geoscientific Model Development},
  volume={9},
  number={5},
  pages={1937--1958},
  year={2016},
  publisher={Copernicus GmbH}
}

@article{govett2024exascale,
  title={Exascale computing and data handling: Challenges and opportunities for weather and climate prediction},
  author={Govett, Mark and Bah, Bubacar and Bauer, Peter and Berod, Dominique and Bouchet, Veronique and Corti, Susanna and Davis, Chris and Duan, Yihong and Graham, Tim and Honda, Yuki and others},
  journal={Bulletin of the American Meteorological Society},
  volume={105},
  number={12},
  pages={E2385--E2404},
  year={2024},
  publisher={American Meteorological Society}
}

@article{kay2015community,
  title={The Community Earth System Model (CESM) large ensemble project: A community resource for studying climate change in the presence of internal climate variability},
  author={Kay, Jennifer E and Deser, Clara and Phillips, A and Mai, A and Hannay, Cecile and Strand, Gary and Arblaster, Julie Michelle and Bates, SC and Danabasoglu, Gokhan and Edwards, James and others},
  journal={Bulletin of the American Meteorological Society},
  volume={96},
  number={8},
  pages={1333--1349},
  year={2015}
}

@article{watt2025ace2,
  title={ACE2: accurately learning subseasonal to decadal atmospheric variability and forced responses},
  author={Watt-Meyer, Oliver and Henn, Brian and McGibbon, Jeremy and Clark, Spencer K and Kwa, Anna and Perkins, W Andre and Wu, Elynn and Harris, Lucas and Bretherton, Christopher S},
  journal={npj Climate and Atmospheric Science},
  volume={8},
  number={1},
  pages={205},
  year={2025},
  publisher={Nature Publishing Group UK London}
}

@article{kochkov2024neural,
  title={Neural general circulation models for weather and climate},
  author={Kochkov, Dmitrii and Yuval, Janni and Langmore, Ian and Norgaard, Peter and Smith, Jamie and Mooers, Griffin and Kl{\"o}wer, Milan and Lottes, James and Rasp, Stephan and D{\"u}ben, Peter and others},
  journal={Nature},
  volume={632},
  number={8027},
  pages={1060--1066},
  year={2024},
  publisher={Nature Publishing Group UK London}
}

@inproceedings{liu2024st,
  title={St-llm: Large language models are effective temporal learners},
  author={Liu, Ruyang and Li, Chen and Tang, Haoran and Ge, Yixiao and Shan, Ying and Li, Ge},
  booktitle={European Conference on Computer Vision},
  pages={1--18},
  year={2024},
  organization={Springer}
}

@article{polyak2024movie,
  title={Movie gen: A cast of media foundation models},
  author={Polyak, Adam and Zohar, Amit and Brown, Andrew and Tjandra, Andros and Sinha, Animesh and Lee, Ann and Vyas, Apoorv and Shi, Bowen and Ma, Chih-Yao and Chuang, Ching-Yao and others},
  journal={arXiv preprint arXiv:2410.13720},
  year={2024}
}

@article{hacohen2024ltx,
  title={Ltx-video: Realtime video latent diffusion},
  author={HaCohen, Yoav and Chiprut, Nisan and Brazowski, Benny and Shalem, Daniel and Moshe, Dudu and Richardson, Eitan and Levin, Eran and Shiran, Guy and Zabari, Nir and Gordon, Ori and others},
  journal={arXiv preprint arXiv:2501.00103},
  year={2024}
}

@article{kong2024hunyuanvideo,
  title={Hunyuanvideo: A systematic framework for large video generative models},
  author={Kong, Weijie and Tian, Qi and Zhang, Zijian and Min, Rox and Dai, Zuozhuo and Zhou, Jin and Xiong, Jiangfeng and Li, Xin and Wu, Bo and Zhang, Jianwei and others},
  journal={arXiv preprint arXiv:2412.03603},
  year={2024}
}

@article{kaltenborn2023climateset,
  title={Climateset: A large-scale climate model dataset for machine learning},
  author={Kaltenborn, Julia and Lange, Charlotte and Ramesh, Venkatesh and Brouillard, Philippe and Gurwicz, Yaniv and Nagda, Chandni and Runge, Jakob and Nowack, Peer and Rolnick, David},
  journal={Advances in Neural Information Processing Systems},
  volume={36},
  pages={21757--21792},
  year={2023}
}

@article{hersbach2020era5,
  title={The ERA5 global reanalysis},
  author={Hersbach, Hans and Bell, Bill and Berrisford, Paul and Hirahara, Shoji and Hor{\'a}nyi, Andr{\'a}s and Mu{\~n}oz-Sabater, Joaqu{\'\i}n and Nicolas, Julien and Peubey, Carole and Radu, Raluca and Schepers, Dinand and others},
  journal={Quarterly journal of the royal meteorological society},
  volume={146},
  number={730},
  pages={1999--2049},
  year={2020},
  publisher={Wiley Online Library}
}

@article{nguyen2023climax,
  title={Climax: A foundation model for weather and climate},
  author={Nguyen, Tung and Brandstetter, Johannes and Kapoor, Ashish and Gupta, Jayesh K and Grover, Aditya},
  journal={arXiv preprint arXiv:2301.10343},
  year={2023}
}

@inproceedings{ronneberger2015u,
  title={U-net: Convolutional networks for biomedical image segmentation},
  author={Ronneberger, Olaf and Fischer, Philipp and Brox, Thomas},
  booktitle={International Conference on Medical image computing and computer-assisted intervention},
  pages={234--241},
  year={2015},
  organization={Springer}
}

@article{chen2024diffusion,
  title={Diffusion forcing: Next-token prediction meets full-sequence diffusion},
  author={Chen, Boyuan and Mart{\'\i} Mons{\'o}, Diego and Du, Yilun and Simchowitz, Max and Tedrake, Russ and Sitzmann, Vincent},
  journal={Advances in Neural Information Processing Systems},
  volume={37},
  pages={24081--24125},
  year={2024}
}

@article{ho2022video,
  title={Video diffusion models},
  author={Ho, Jonathan and Salimans, Tim and Gritsenko, Alexey and Chan, William and Norouzi, Mohammad and Fleet, David J},
  journal={Advances in neural information processing systems},
  volume={35},
  pages={8633--8646},
  year={2022}
}

@article{bouabid2024fairgp,
  title={FaIRGP: A Bayesian energy balance model for surface temperatures emulation},
  author={Bouabid, Shahine and Sejdinovic, Dino and Watson-Parris, Duncan},
  journal={Journal of Advances in Modeling Earth Systems},
  volume={16},
  number={6},
  pages={e2023MS003926},
  year={2024},
  publisher={Wiley Online Library}
}

@inproceedings{esser2024scaling,
  title={Scaling rectified flow transformers for high-resolution image synthesis},
  author={Esser, Patrick and Kulal, Sumith and Blattmann, Andreas and Entezari, Rahim and M{\"u}ller, Jonas and Saini, Harry and Levi, Yam and Lorenz, Dominik and Sauer, Axel and Boesel, Frederic and others},
  booktitle={Forty-first international conference on machine learning},
  year={2024}
}

@article{vaswani2017attention,
  title={Attention is all you need},
  author={Vaswani, Ashish and Shazeer, Noam and Parmar, Niki and Uszkoreit, Jakob and Jones, Llion and Gomez, Aidan N and Kaiser, {\L}ukasz and Polosukhin, Illia},
  journal={Advances in neural information processing systems},
  volume={30},
  year={2017}
}

@article{dehghani2023patch,
  title={Patch n’pack: Navit, a vision transformer for any aspect ratio and resolution},
  author={Dehghani, Mostafa and Mustafa, Basil and Djolonga, Josip and Heek, Jonathan and Minderer, Matthias and Caron, Mathilde and Steiner, Andreas and Puigcerver, Joan and Geirhos, Robert and Alabdulmohsin, Ibrahim M and others},
  journal={Advances in Neural Information Processing Systems},
  volume={36},
  pages={2252--2274},
  year={2023}
}

@article{su2024roformer,
  title={Roformer: Enhanced transformer with rotary position embedding},
  author={Su, Jianlin and Ahmed, Murtadha and Lu, Yu and Pan, Shengfeng and Bo, Wen and Liu, Yunfeng},
  journal={Neurocomputing},
  volume={568},
  pages={127063},
  year={2024},
  publisher={Elsevier}
}

@article{tebaldi2025emulators,
  title={Emulators of climate model output},
  author={Tebaldi, Claudia and Selin, NE and Ferrari, R and Flierl, G},
  journal={Annual Review of Environment and Resources},
  volume={50},
  year={2025},
  publisher={Annual Reviews}
}

@techreport{forster2021emulators,
  title={The Role of Climate Model Emulators in IPCC AR6},
  author={Forster, Piers M. and Jackson, Luke S. and Smith, Chris J. and Rogelj, Joeri and others},
  year={2021},
  institution={CONSTRAIN EU Project},
  url={https://www.constrain-eu.org/wp-content/uploads/2021/10/The_Role_of_Climate_Model_Emulators_in_IPCC_AR6_D4.3.pdf},
  note={Used for global mean temperature and sea level projections at annual timescales}
}

@article{niu2024multi,
  title={Multi-fidelity residual neural processes for scalable surrogate modeling},
  author={Niu, Ruijia and Wu, Dongxia and Kim, Kai and Ma, Yi-An and Watson-Parris, Duncan and Yu, Rose},
  journal={arXiv preprint arXiv:2402.18846},
  year={2024}
}

@article{lutjens2025impact,
  title={The impact of internal variability on benchmarking deep learning climate emulators},
  author={L{\"u}tjens, Bj{\"o}rn and Ferrari, Raffaele and Watson-Parris, Duncan and Selin, Noelle E},
  journal={Journal of Advances in Modeling Earth Systems},
  volume={17},
  number={8},
  pages={e2024MS004619},
  year={2025},
  publisher={Wiley Online Library}
}

@article{rasp2020weatherbench,
  title={WeatherBench: a benchmark data set for data-driven weather forecasting},
  author={Rasp, Stephan and Dueben, Peter D and Scher, Sebastian and Weyn, Jonathan A and Mouatadid, Soukayna and Thuerey, Nils},
  journal={Journal of Advances in Modeling Earth Systems},
  volume={12},
  number={11},
  pages={e2020MS002203},
  year={2020},
  publisher={Wiley Online Library}
}

@article{matheson1976scoring,
  title={Scoring rules for continuous probability distributions},
  author={Matheson, James E and Winkler, Robert L},
  journal={Management science},
  volume={22},
  number={10},
  pages={1087--1096},
  year={1976},
  publisher={INFORMS}
}

@article{bassetti2024diffesm,
  title={{DiffESM}: Conditional emulation of temperature and precipitation in Earth System Models with 3D diffusion models},
  author={Bassetti, Seth and Hutchinson, Brian and Tebaldi, Claudia and Kravitz, Ben},
  journal={Journal of Advances in Modeling Earth Systems},
  volume={16},
  number={10},
  year={2024},
  pages={e2023MS004194}
}

@article{brenowitz2025climate,
  title={Climate in a Bottle: Towards a Generative Foundation Model for the Kilometer-Scale Global Atmosphere},
  author={Brenowitz, Noah D. and Ge, Tao and Subramaniam, Akshay and Manshausen, Peter and Gupta, Aayush and Hall, David M. and Mardani, Morteza and Vahdat, Arash and Kashinath, Karthik and Pritchard, Michael S.},
  journal={arXiv preprint arXiv:2505.06474},
  year={2025}
}

@inproceedings{srivastava2024precip,
  title={Precipitation Downscaling with Spatiotemporal Video Diffusion},
  author={Srivastava, Prakhar and Yang, Ruihan and Kerrigan, Gavin and Dresdner, Gideon and McGibbon, Jeremy},
  booktitle={Advances in Neural Information Processing Systems (NeurIPS)},
  year={2024}
}

@article{dinh2016density,
  title={Density estimation using real nvp},
  author={Dinh, Laurent and Sohl-Dickstein, Jascha and Bengio, Samy},
  journal={arXiv preprint arXiv:1605.08803},
  year={2016}
}

@article{kingma2018glow,
  title={Glow: Generative flow with invertible 1x1 convolutions},
  author={Kingma, Durk P and Dhariwal, Prafulla},
  journal={Advances in neural information processing systems},
  volume={31},
  year={2018}
}

@article{saharia2022image,
  title={Image super-resolution via iterative refinement},
  author={Saharia, Chitwan and Ho, Jonathan and Chan, William and Salimans, Tim and Fleet, David J and Norouzi, Mohammad},
  journal={IEEE transactions on pattern analysis and machine intelligence},
  volume={45},
  number={4},
  pages={4713--4726},
  year={2022},
  publisher={IEEE}
}

@article{dhariwal2021diffusion,
  title={Diffusion models beat gans on image synthesis},
  author={Dhariwal, Prafulla and Nichol, Alexander},
  journal={Advances in neural information processing systems},
  volume={34},
  pages={8780--8794},
  year={2021}
}

@article{zhou2022magicvideo,
  title={Magicvideo: Efficient video generation with latent diffusion models},
  author={Zhou, Daquan and Wang, Weimin and Yan, Hanshu and Lv, Weiwei and Zhu, Yizhe and Feng, Jiashi},
  journal={arXiv preprint arXiv:2211.11018},
  year={2022}
}

@misc{MetOffice_arise-cmor-tables_2025,
  author       = {{Met Office}},
  title        = {{UKESM1‐0 ARISE‐SAI information and CMOR3 tables (arise-cmor-tables)}},
  howpublished = {\url{https://github.com/MetOffice/arise-cmor-tables}},
  year         = {2025},
  note         = {Accessed: 2025-11-12}
}

@software{Andela_ESMValCore_2025,
author = {Andela, Bouwe and Broetz, Bjoern and de Mora, Lee and Drost, Niels and Eyring, Veronika and Koldunov, Nikolay and Lauer, Axel and Predoi, Valeriu and Righi, Mattia and Schlund, Manuel and Vegas-Regidor, Javier and Zimmermann, Klaus and Bock, Lisa and Diblen, Faruk and Dreyer, Laura and Earnshaw, Paul and Hassler, Birgit and Little, Bill and Loosveldt-Tomas, Saskia and Smeets, Stef and Camphuijsen, Jaro and Gier, Bettina K. and Weigel, Katja and Hauser, Mathias and Kalverla, Peter and Galytska, Evgenia and Cos-Espuña, Pep and Pelupessy, Inti and Koirala, Sujan and Stacke, Tobias and Alidoost, Sarah and Jury, Martin and Sénési, Stéphane and Crocker, Thomas and Vreede, Barbara and Soares Siqueira, Abel and Kazeroni, Rémi and Hohn, David and Bauer, Julian and Beucher, Romain and Benke, Joerg and Martin-Martinez, Eneko and Cammarano, Diego and Yousong, Zeng and Malinina, Elizaveta and Garcia Perdomo, Karen and Lenhardt, Julien},
doi = {10.5281/zenodo.3387139},
license = {Apache-2.0},
month = oct,
title = {{ESMValCore}},
url = {https://github.com/ESMValGroup/ESMValCore/},
version = {v2.13.0},
year = {2025}
}

@software{Hassan_acccmip6_Python_package_2022,
author = {Hassan, Taufiq},
doi = {10.5281/zenodo.6559056},
month = may,
title = {{acccmip6: Python package for accessing and downloading CMIP6 data}},
url = {https://github.com/TaufiqHassan/acccmip6},
version = {5.2.0},
year = {2022}
}

@article{watt2024generative,
  title={Generative diffusion-based downscaling for climate},
  author={Watt, Robbie A and Mansfield, Laura A},
  journal={arXiv preprint arXiv:2404.17752},
  year={2024}
}

@article{ling2024diffusion,
  title={Diffusion model-based probabilistic downscaling for 180-year East Asian climate reconstruction},
  author={Ling, Fenghua and Lu, Zeyu and Luo, Jing-Jia and Bai, Lei and Behera, Swadhin K and Jin, Dachao and Pan, Baoxiang and Jiang, Huidong and Yamagata, Toshio},
  journal={npj Climate and Atmospheric Science},
  volume={7},
  number={1},
  pages={131},
  year={2024},
  publisher={Nature Publishing Group UK London}
}
